\documentclass{article}

\usepackage{microtype}
\usepackage{graphicx}
\usepackage{subfigure}
\usepackage{booktabs} 

\usepackage{hyperref}

\usepackage[accepted]{icml2025}

\usepackage{amsmath}
\usepackage{amssymb}
\usepackage{mathtools}
\usepackage{amsthm}

\usepackage[capitalize,noabbrev]{cleveref}

\usepackage{wrapfig}
\usepackage{caption}
\usepackage{amsmath}
\usepackage{imakeidx}
\makeindex[columns=3, title=Alphabetical Index, intoc]
\usepackage{longtable}
\usepackage{minitoc}
\usepackage{listings}
\usepackage{xcolor}

\definecolor{codegreen}{rgb}{0,0.6,0}
\definecolor{codegray}{rgb}{0.5,0.5,0.5}
\definecolor{codepurple}{rgb}{0.58,0,0.82}
\definecolor{backcolour}{rgb}{0.95,0.95,0.92}
\lstdefinestyle{mystyle}{
    backgroundcolor=\color{backcolour},   
    commentstyle=\color{codegreen},
    keywordstyle=\color{magenta},
    numberstyle=\tiny\color{codegray},
    stringstyle=\color{codepurple},
    basicstyle=\ttfamily\footnotesize,
    breakatwhitespace=false,         
    breaklines=true,                 
    captionpos=b,                    
    keepspaces=true,                 
    numbers=left,                    
    numbersep=5pt,                  
    showspaces=false,                
    showstringspaces=false,
    showtabs=false,                  
    tabsize=2
}
\usepackage[utf8]{inputenc} 
\usepackage[T1]{fontenc}    
\usepackage{hyperref}       
\usepackage{url}            
\usepackage{booktabs}       
\usepackage{amsfonts}       
\usepackage{nicefrac}       
\usepackage{microtype}      
\usepackage{xcolor}         
\usepackage{makecell} 
\usepackage{cuted}
\usepackage{array}
\usepackage{multirow}
\usepackage{enumitem}
\usepackage{rotating}
\newcommand{\quotes}[1]{``#1''}

\theoremstyle{plain}

\theoremstyle{definition}

\theoremstyle{remark}

\usepackage[textsize=tiny]{todonotes}
\icmltitlerunning{Neuro-Symbolic Programmatic Representations for Inductive Spatial Concepts}

\begin{document}

\twocolumn[
\icmltitle{\centerline{\emph{Sketch-Plan-Generalize} : Learning and Planning with} Neuro-Symbolic Programmatic Representations for Inductive Spatial Concepts}
%
%
%
\icmlsetsymbol{equal}{*}
%
\begin{icmlauthorlist}
\icmlauthor{Namasivayam Kalithasan}{equal,sch}
\icmlauthor{Sachit Sachdeva}{equal,sch}
\icmlauthor{Himanshu Gaurav Singh}{when}
\icmlauthor{Vishal Bindal}{when}
\icmlauthor{Arnav Tuli}{when}
\icmlauthor{Gurarmaan Singh Panjeta}{sch}
\icmlauthor{Harsh Himanshu Vora}{sch}
\icmlauthor{Divyanshu Aggarwal}{sch}
\icmlauthor{Rohan Paul}{sch}
\icmlauthor{Parag Singla}{sch}
\end{icmlauthorlist}
\icmlaffiliation{when}{Work Done when at IIT Delhi}
\icmlaffiliation{sch}{IIT Delhi}
%
%
\icmlcorrespondingauthor{Namasivayam Kalithasan}{Namasivayam.k@cse.iitd.ac.in}
%
\icmlkeywords{Machine Learning, ICML}
\vskip 0.3in 
]



\printAffiliationsAndNotice{\icmlEqualContribution} 

\begin{abstract}
Effective human-robot collaboration requires the ability to learn personalized concepts from a limited number of demonstrations, while exhibiting inductive generalization, hierarchical composition, and adaptability to novel constraints. 
Existing approaches that use code generation capabilities of pre-trained large (vision) language models as well as purely neural models show poor generalization to \emph{a-priori} unseen complex concepts.  Neuro-symbolic methods~\cite{grand2023lilo} offer a promising alternative by searching in program space, but face challenges in large program spaces due to the inability to effectively guide the search using demonstrations.
Our key insight is to factor inductive concept learning as: (i) {\it Sketch:} detecting and inferring a coarse signature of a new concept (ii) {\it Plan:} performing an MCTS search over grounded action sequences guided by human demonstrations (iii) {\it Generalize:} abstracting out  grounded plans as inductive programs. Our pipeline facilitates generalization and modular re-use, enabling continual concept learning.  
Our approach combines the benefits of code generation ability of large language models (LLMs) along with grounded neural representations, resulting in neuro-symbolic programs that show stronger inductive generalization on the task of constructing complex structures vis-\'a-vis LLM-only and purely neural approaches. Further, we demonstrate reasoning and planning capabilities with learned concepts for embodied instruction following. 
\end{abstract}
\section{Introduction}

For robots to collaborate effectively with humans, they must quickly learn personalised concepts from just a few demonstrations. This ability to form inductive representations of novel grounded concepts and apply them for tasks beyond training is a hallmark of human intelligence~\citep{mind, chollet2019measureARC-AGI}. In the context of robot learning, such representations are characterized by four fundamental properties: (a) it should be grounded directly in human demonstrations, reflecting the human’s intent rather than relying sorely on prior world knowledge, (b) it must generalize beyond observed examples; e.g., from a few towers of certain heights, infer how to build towers of arbitrary heights,
(c) it should allow for the hierarchical interpretation of increasingly complex concepts as compositions of simpler ones; e.g., a staircase as a sequence of towers of increasing height, (d) it should be easily modifiable to incorporate additional constraints in new instructions, such as ensuring no green block is placed near a blue or yellow one in a staircase.


Existing approaches fall short of fully satisfying these criteria. Recent advances in large language models (LLMs) have enabled general-purpose planners. These planners use world knowledge to identify relevant concepts and generate plans from language task specifications~\citep{singh2022progprompt, code-as-policies, ahn2022saycan, achiam2023gpt, Huang2022LanguageMA}. However, they often struggle to learn new concepts from demonstrations, especially when the concepts are linguistically novel. This is because they rely heavily on prior world knowledge, which limits their faithfulness to demonstrations and their ability to generalize inductively. Purely neural approaches, such as~\citep{struct-diffusion}, can learn directly from demonstrations but typically fail to generalize beyond training data because they (a) do not explicitly model symbolic induction and (b) lack modularity, restricting their ability to reuse learned concepts. In contrast, neuro-symbolic approaches~\citep{grand2023lilo, ellis2021dreamcoder} offer a more promising alternative. They leverage LLMs or enumerative program search to discover generalizable programs directly in the program (string) space. While these models outperform purely LLM-based or neural methods in terms of generalization, they still face significant challenges: enumerative searches become computationally infeasible in complex program spaces (e.g., Python programs), and such search cannot be easily guided by human demonstrations.

\begin{figure*}[h]
   \centering
   \includegraphics[width=0.9\textwidth]{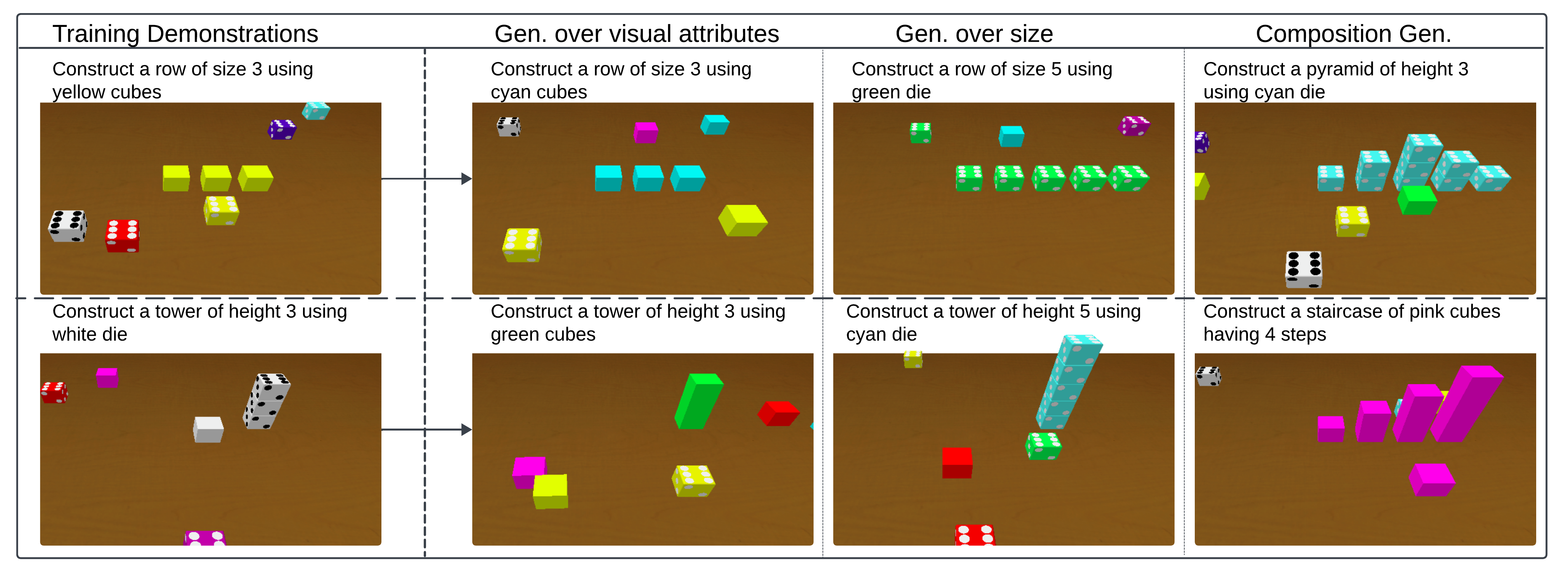}
   \caption{{\footnotesize{\textbf{Problem Overview.} Our goal is to enable an embodied agent to learn grounded and generalizeable representations for spatial abstractions possessing a notion of induction (e.g., constructing a tower, row or their combinations such as staircases, boundary etc.). (Left) A human demonstrates the construction of a \emph{row} and \emph{tower} of size three. (Right) The agent learns \emph{program} representation that enables  inductive generalization to novel structures (varied sizes and visual attributes) and expresses complex concepts as hierarchical composition of previously acquired ones. E.g., \emph{tower} as blocks placed one on top of another and a  \emph{pyramid} as rows of decreasing size.}}}
   \label{fig:generalization}
    \vspace*{-0.3cm}
\end{figure*}


%
%
In response, we present a neuro-symbolic agentic framework that (i) identifies when learning a new concept is required, (ii) requests human demonstrations to learn an inductively generalizable representation of the unknown concept, and (iii) uses the learned concepts to generate control policies for novel instruction-following tasks with discrete constraints. Unlike prior methods, our concept learning approach, called \emph{SPG}, \emph{factorizes} the concept learning task into three stages: (a) \emph{\underline{S}ketch}: Given a language-annotated demonstration of a novel concept, an LLM is used to postulate a function signature. (b) \emph{\underline{P}lan}: Using MCTS, grounded action sequences are discovered by maximizing a reward that measures how closely the constructed structure matches the human demonstration. (c) \emph{\underline{G}eneralize}: The code-generation capability of an LLM is leveraged to distill grounded plans from multiple demonstrations into an inductively generalizable program. This enables learning of inductively generalizable representations while restricting the search to the grounded actions, which can be effectively guided through dense rewards from human demonstration and is less expensive than searching directly in the program space. We also show how the learned concepts can be used for goal-conditioned planning in novel input scenes and instructions, even when visual or spatial constraints are present. Our experiments demonstrate that our framework can accurately learn both simple and complex concepts from just a few demonstrations across various spatial structures. Moreover, it exhibits strong inductive generalization in out-of-distribution settings, significantly outperforming baseline methods.  

\section{Related Works} \label{sec:related}
\textbf{Concept Learning:} 
The problem of acquiring higher-order programmatic constructs is often modelled as Bayesian inference over a latent symbol space given observed instances. Seminal works have demonstrated efficient inference over latent generative programs to express handwritten digits~\citep{lake2015human}, 2D drawings~\citep{ellis2018learning}, grid-world puzzles~\citep{wang2024hypothesis}, visual question answering~\citep{nscl}, goal-directed policies~\citep{lpp}, and compressed or refactored code~\citep{grand2023lilo, ellis2021dreamcoder}. These approaches typically either (i) search program space using a probabilistic grammar, which is infeasible for large program spaces like Python, or (ii) use LLMs to directly infer programs, which may produce physically ungrounded outputs. Furthermore, because partially generated programs cannot be evaluated, direct search in program space does not effectively leverage demonstrations to guide the search.
In contrast, our approach first searches for grounded action sequences that can be efficiently guided by dense rewards from human demonstrations. We then use an LLM to abstract these sequences into a program, making program inference more tractable and efficient. While prior works focus on 2D problems without considering physical plausibility (for example, towers can be drawn from top to bottom but cannot be constructed that way in reality), we expose the search to assessing the physical construction plausibility, thereby learning physically grounded concepts. 
%

\textbf{Learning-to-plan Methods:} Our work is complementary to efforts that learn symbolic constructs for efficient planning. Works such as~\cite{silver2023predicate,silver2024generalized,liu2024learning}, infer state-action abstractions for planning by querying large pre-trained models or by optimizing a goal attainability objective. This paper, instead, focuses on learning a representation for complex spatial assemblies as inductive programs leading to the ability to infer complex goal specifications which can then be combined with aforementioned works for synthesize efficient plans to realize complex assemblies.  Works such as \cite{li2019practical} learn to construct structures by encoding relational knowledge via graph neural network. However, this effort suffers from poor generalization to unseen examples, (e.g., tower of larger size) and do not possess a mechanism to re-use previously acquired concepts. Works such as \cite{wang2023voyager,jarvis-1} shows lifelong learning of skills by learning to plan high-level tasks through composition of simple skills for simulated agents. Others~\citep{lotus,lifelong-halp} initiate new skill acquisition upon detecting task failure, building a library of skills over time. However, they do not model inductive use of learned concepts and initiate skill acquisition only upon failure as opposed to learning continually even from goal-reaching demonstrations.   

\textbf{Robot Instruction Following:} Instruction following involves grounding symbolic constructs expressed in language with aspects of the state-action space such as object assemblies~\citep{paul2018efficient,ramp,lachmy2022draw},  spatial relations  ~\citep{tellex2011approaching,kim2024lingo}, reward functions  ~\citep{boularias2015grounding}, or motion constraints ~\citep{howard2014natural}. These works assume the presence of grounded representation for symbolic concepts and only learn associations between language and concepts. In contrast our work \emph{jointly} learns higher-order concepts composed of simpler concepts along with their grounding in the robot's state and action space. Others~\citep{singh2022progprompt,wang2023demo2code,ahn2022saycan,code-as-policies} leverage prior-knowledge embodied in large vision-language models to directly translate high-level tasks to robot control programs. Our experiments (reported subsequently) demonstrate their limitation in outputting programs for structure assembly-type tasks that require long-range (inductive) spatial reasoning and consideration of physical plausability of construction. Our approach addresses this problem by \emph{coupling}  abstract task knowledge from pre-trained models with physical reasoning in the space of executable plans. 

\section{Preliminaries and Problem Setting} 
\label{sec:background}

We consider an embodied agent that uses a visual and depth sensor to observe its environment and can grasp and release objects at specified poses. We represent the robot's domain as a goal-conditioned MDP $<\mathcal{S}, \mathcal{A}, \mathcal{T}, g, \mathcal{R}, \gamma>$ where $\mathcal{S}$ is the state space, $\mathcal{A}$ is the action space, $\mathcal{T}$ is the transition function, $g$ is the goal, $R$ is the reward model and $\gamma$ is the discount factor. The agent's objective is to learn a policy that generates a sequence of actions from an initial state $s_0$ to achieve the goal $g$ in response to an instruction $\Lambda$ specifying the intended goal from a human.  
We assume that the agent possesses a model of semantic relations (e.g., left(), right() etc.) as well as semantic actions such as moving an object by grasping and releasing at a target location. Such modular and composable notions can be acquired from demonstrations via approaches outlined in~\cite{nsrm-Kalithasan2022LearningNP, nscl, mao2022pdsketch}. Such notions populate a library of concepts $\mathcal{L}$ available as grounded executable function calls. Following recent efforts~\citep{code-as-policies,Huang2022InnerME,ahn2022saycan,singh2022progprompt} in representing robot control directly as executable programs, we represent action sequence corresponding to a plan as a program consisting of function calls to executable actions and grounded spatial reasoning.  

Our goal is to enable a robot to interpret and learn the concepts in instructions such as \emph{``construct a tower with red blocks of height five"}. Specifically, we aim to learn spatial constructs like a tower that requires sequential actions that repeatedly place a block on top of a previously constructed assembly, a process akin to induction. Given a few demonstrations of constructing a spatial assembly, $\mathcal{D}$, each consisting of natural language description $\Lambda$ (``construct a tower of red blocks of size five") and a sequence of key frame states $\{S_1, \cdots, S_g\}$ associated with the construction process, we seek to learn a program that models the inductive nature of the concept of tower. This learned representation should enable the agent to generalize inductively to new instructions, such as "construct a tower of blue blocks of height ten." Moreover, the learned concepts should facilitate the learning of more complex structures, which are challenging to represent using primitive actions alone. For example, the concept of a "tower" should assist in learning a "staircase," which can be represented as a sequence of towers of increasing heights.

\section{Representing Inductive Spatial Concepts}  \label{sec:representation}
%
We formalize the notion of inductive spatial concepts and formulate the learning objective.

\textbf{Inductive Spatial Concepts:}  A spatial structure is an inductive concept if its construction can be described recursively using a similar structure of smaller size or as a composition of other simpler structures. Formally, let $C_1, \cdots, C_{|\mathcal{L}|}$ represent the concepts in the concept library $\mathcal{L}$. We define a partial order on $\mathcal{L}$ where a concept $C$ is "dependent on" $\widetilde{C}$ if $\widetilde{C}$ is a substructure of $C$. For example, a staircase is dependent on a tower, and $X$ (cross) is dependent on diagonals, and so on. This partial order is referred to as structural complexity, where a concept $C$ is more structurally complex than $\widetilde{C}$ if $C$ is dependent on $\widetilde{C}$. Without loss of generality, assume that $C_1, \cdots, C_{|\mathcal{L}|}$ are written in topological order as per their structural complexity. Now, the construction of an inductive spatial concept $C_k$ of size $n$ at position $p$, denoted by $h(C_k, n, p)$, is defined recursively as:
\begin{multline}
     h(C_k, n, p) = \underbrace{h ^ \lambda (C_k, n-1, pos(.))}_{\text{Induction (I)}} \circ \\
     \underbrace{\prod\limits_{l=1}^{L(C_k)} h\big(C_{k^\prime_l}, \mathtt{size}(.), pos(.)\big)}_{\text{Composition (C)}} \circ 
     \underbrace{\prod\limits_{l=1}^{L^\prime(C_k)} \eta_{\theta}^{l}\big( pos(.)\big)}_{\text{Base (B)}}
    \label{eqn:inductive_cpt_def}
\end{multline}
where, $\lambda\in\{0,1\}$, $k^\prime < k$, $0\leq L(C_k), L^\prime(C_k) \leq o(|\mathcal{L}|), pos(.) = pos(C_k, l,n,p)$ and $\mathtt{size}(.) = \mathtt{size}(C_k, l, n)$ are functions that predict the size and position of the structure to be constructed.
\begin{enumerate}
    \item \textit{Induction term:} The first expression (I) represents induction vis-\'a-vis the possibility of constructing $C_k$ of size $n$ using $C_k$ of size $n-1$. Here, $\lambda$ depicts the absence($\lambda=0$) or presence($\lambda=1$) of the (I) term.
    \item \textit{Composition term:} The second term (C)  expresses the construction of $C_k$ as a composition of prev. known concepts in the library ($\mathcal{L}$). The number of required compositions ($L(C_k)$) depends on $C_k$ and $\mathtt{len}(\mathcal{L})$. 
    \item \textit{Base term:} The third term (B) defines the base case where the construction of concept $C_k$ may include $L^\prime$ number of primitive actions. Here, $\eta_{\theta}$ are pre-trained primitive actions resulting in the repositioning of an object to the desired spatial position. For ex., a tower of size $n$ can be constructed by a (primitive) move-top action after constructing a tower of size $n-1$.
\end{enumerate} 
%
%
%

\textbf{Learning Objective:} The functional space of inductive concepts ($h$) leads to a hypothesis space $\mathcal{H}$ of associated neuro-symbolic programs.
Each goal-reaching demonstration corresponds to a particular instantiation of a given inductive concept, i.e. $h(C_{k}, n, p)$, where the $p$ comes from the sequence of frames, and $n$, $C_{k}$ comes from $\Lambda$. We aim to learn a generic representation $H = h(C_{k}, \cdot, \cdot) \in \mathcal{H}$ for the given concept, which is general for all $n$ and $p$. Given (few) demonstrations of a human constructing a spatial structure, concept learning can be formulated as the Bayesian posterior~\citep{lake2015human,shah2018bayesian,lpp}, $\mathtt{P}_\mathcal{H}(H | \Lambda, S_{1}..S_{g}) \propto \mathtt{P}(S_{1}..S_{g} | \Lambda, H) \cdot \mathtt{P}(H | \Lambda)$. Here, the likelihood term associates a candidate program, and the prior term regularizes the program space. The maximum \emph{a-posteriori} estimate, representing the learnt program, is obtained by optimizing the following objective: 
\begin{multline}
   H^* = \arg \underset{H \in \mathcal{H}}{\mathrm{min}} [ \mathtt{Loss}(\{S_{1}..S_{g}\}, Exec(H, \Lambda, S_{1})) \\ 
   - \log \mathtt{P}(H | \Lambda) ]
    \label{eqn:learning-objective}
\end{multline}
Since exact inference is intractable, approximate inference is performed via search in the program space. Note that learning inductive spatial concepts given demonstration considers programs that represent plans that attain physically grounded/feasible structures, an object we model during the search. Additionally, we seek strong generalization from a few instances of an inductive structure to structures with arbitrary sizes, in effect favouring programs with iterative looping constructs.


\section{Learning Inductive Concepts from Demonstrations}
\begin{figure*}[h]
    \centering
    \includegraphics[width=0.9\textwidth, height=6cm]{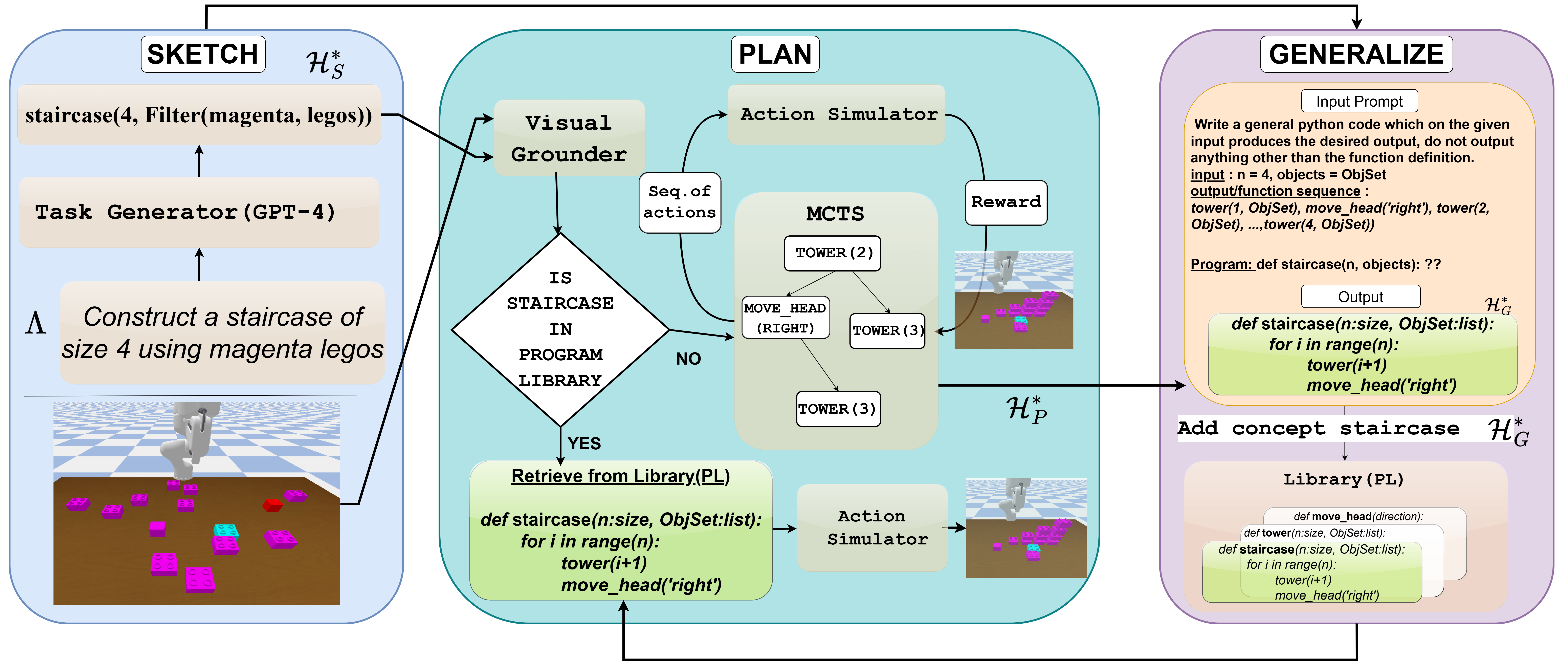}
    \caption{
    \footnotesize{\textbf{Method Overview.} We learn a neuro-symbolic program for inductive spatial concepts factored as (a) \emph{Sketch} (b) \emph{Plan} (c) \emph{Generalize}. The example above shows the progressive realization of a program for the concept of a staircase acquired by observing a single human demonstration of building a staircase of size four, and its corresponding language instruction}}
    \label{fig:overall}
    \vspace*{-0.3cm}
\end{figure*}

For estimating a succinct generalized program as per Eq. ~\ref{eqn:learning-objective}, a direct symbolic search in the space of programs can explicitly reason over previously acquired concepts, it is intractable, particularly due to looping constructs needed for modeling induction. Alternatively, neural methods that attempt to directly predict action sequence to explain the demonstrations are resilient to noise, but are challenged in the continual setting as the number of concepts can increase prohibitively over time. Our approach blends both approaches and factors the concept learning task as:
\begin{itemize}
\item \textbf{Sketch:} From the natural language instruction ($\Lambda$), we extract a task sketch ($H_S^*$) using an LLM that provides the signature (concept name and instantiated arguments) of the concept to be learned. \item \textbf{Plan:} An MCTS-based search is performed using the library concepts (learnt previously) which outputs a sequence of grounded actions that best explain the given demonstration. \item \textbf{Generalize:} The grounded plan $H_P^*$ and task sketch $H_S^*$ are provided to an LLM to obtain a general Python program whose execution on the given scene matches the searched plan.
\end{itemize}
Formally, the factored exploration of the program space for a demonstration is performed as: 
\begin{multline}
H_S^* \leftarrow Sketch(\Lambda \, ; \,{\theta_S}) \, ; \, H_P^* \leftarrow Plan(S_{1}..S_{g}, H_S^*; \,{\theta_P}) \, ;\\ \,
 H_G^* \leftarrow Generalize(H_P^*, \, H_S^*; \,{\theta_G})
 \end{multline}
Here, $\theta_S$, $\theta_P$ and $\theta_G$ are the learnable parameters (including hyperparameters) of the Sketch, Plan and Generalize functions, respectively. The concept library $\mathcal{L}$ is initialized with primitive visual and action concepts. Upon acquiring a new inductive concept $H_G^* = H^*$, we update our library accordingly: $\mathcal{L} \leftarrow \mathcal{L} \cup {H^*}$. An example is provided in Appendix Sec.~\ref{sec:exp_methodology}.  Fig.~\ref{fig:overall} illustrates an example of progressive prog. estimation. Next, we detail each of these three steps.

\subsection{(Sketch) Grounded Task Sketch Generation}
An LLM driven by in-context learning is used to get a program signature (a sketch) for a concept from the natural language instruction. The task sketch is a tree of nested function calls that outlines the function header (name and the parameters) of the inductive concept/program to be learned. Detailed exposition on prompting appears in the app. ~\ref{app:tsg-prompting}. 

The task sketch thus obtained is then grounded on the input scene using a visual grounding module akin to \cite{nscl, nsrm-Kalithasan2022LearningNP, wang2023progport}. This module has three key components: (1) a visual extractor (ResNet-34 based) that extracts the features of all objects in the scene, (2) a concept embedding module that learns disentangled representations for visual concepts like \emph{green} and \emph{dice}, and (3) an executor equipped with pre-defined behaviours such as \quotes{filter} to select/ground the objects of interest. For example, grounding the task sketch \quotes{Tower (height =3, objects =  filter(green, dice))} results in an instantiated function call \quotes{Tower(height = 3, ObjectSet)} where ObjectList is an ordered list of the green coloured dice. Since it is now grounded in the initial scene of the demonstration, the task sketch corresponds to a particular instance of the concept demonstrated in the given demonstration.

\subsection{(Plan) Physical Reward Guided Plan Search}
\label{subsec:mcts}
The plan search involves finding a generalizable plan that effectively explains the demonstration $S_1,\cdots, S_g$. Specifically, this involves determining the concepts, their respective grounded parameters, and the order of composition as specified in the Equation \ref{eqn:inductive_cpt_def}. 

\textbf{Primitive Actions.} Constructing complex structures involves two steps: (1) identifying or imagining the placement location of an object/structure and (2) picking and placing the object at the imagined location. The position $pos_{\theta}(.)$ for placement is determined using a head, which represents a cuboidal enclosure in 3D space. Conceptually, moving the head is akin to the robot’s cognitive exploration of potential placements to achieve the desired spatial configuration. We define a set of primitive functions, $\mathcal{A}_p$, to guide the movement and placement of objects in two ways: (1) \texttt{move\_head(direction)}: This primitive is a neural operator that moves the abstract head to a desired relative position. It is trained on a corpus of pick-and-place instructions, similar to the approach in \cite{nsrm-Kalithasan2022LearningNP}; and (2) \texttt{keep\_at\_head(objects)}: This primitive places the target object from the list \texttt{objects} at the current location of the head.


\textbf{MCTS Search.} We use an object-centric state representation defined by bounding boxes (including the depth of the center) and visual attributes of all the objects that are present on the table. Each previously learned inductive concept \texttt{<cpt>} has an associated macro-action \texttt{Make\_<cpt>(size)}, that executes the corresponding program for the given size argument, resulting in the construction of the desired concept.  Thus, the set of actions $\mathcal{A}$ is the union of primitive actions $\mathcal{A}_p$ and compound/macro-actions $\mathcal{A}_c$. Intersection over Union (IoU) between the attained state and the expected state in the demonstration is provided as a reward for all macro actions and \texttt{keep\_at\_head(objects)}; all other actions yield zero reward. An MCTS procedure  similar to \cite{complexmctsbackup} (detailed in  Appendix \ref{search-gen-details}) is performed to find a plan (sequence of grounded actions) that maximizes the reward.

\textbf{Modularity and Scalability.} MCTS that searches for a plan only in terms of primitive actions may not be generalizable due to lack of modularity(~\ref{modular-plans}).  The use of macro-actions in the search ensures that the plan $H_p^*$ for a given demonstration is concise, modular, and easily generalizable.  
This can be seen as a form of regularization in terms of the length of concept description by making the prior $P(H) \propto |H|^{-\alpha}$ (where $\alpha$ > 0) in equation (2)
$$
    H^* = \arg \min_{H \in \mathcal{H}} \left[ \mathtt{Loss}(\{S_{1}..S_{n}\}, H(\Lambda)) + \alpha \log |H| \right]
$$
However, as the action space expands with the learning of more concepts, the search becomes slower, necessitating the pruning of the search space. To avoid searching over the size parameter in macro-actions, we greedily select the smallest size that achieves the maximum average reward from the current state. Additionally, to prune primitive actions, we train a reactive policy $\pi_{neural}$ which, given the current state $\tilde{s_{t}}$ and the next expected state ${s_{t+1}}$ (from the demonstration), outputs one of the primitive actions $a_{t}^* \in \mathcal{A}_{p}$. Consequently, the effective branching factor of the search is reduced from $N*|\mathcal{A}_{c}| + |\mathcal{A}_{p}|$ to $|\mathcal{A}_{c}| + 1$, where $N$ is the average number of size-parameter variants. Thus, our MCTS algorithm is modular through the hierarchical composition of learned concepts and efficient through action space pruning, and is referred to as MCTS+\emph{L}+\emph{P}. (Detailed in app. ~\ref{search-gen-details})


\subsection{(Generalize) Plan to Program Abstraction}
Leveraging the code generation and pattern matching abilities of LLMs~\citep{pattern}, we use GPT-4 to distil out a general Python program from the sequence of grounded actions as determined by MCTS+\emph{L}+\emph{P}. The learnt program is incorporated in the concept library, $\mathcal{L}$, for modular reuse in subsequent learning tasks. We take a curriculum learning approach (App. ~\ref{app:continual-neural}). Further details \& discussion can be found in app. items~\ref{app:generalize-prompting},~\ref{app:learning-from-multiple-dem} and ~\ref{search-gen-details}.

\section{Adapting Learned Concepts for Novel Instructions and Constraints}
\label{sec:applications_goal_grounding}
\begin{figure}[h]
    \centering
    \includegraphics[width=0.48\textwidth]{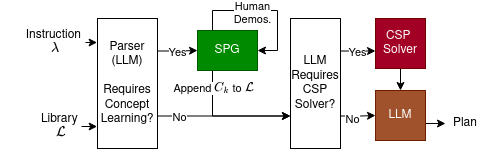}
    \caption{Pipeline for using Solvers and LLMs to solve constraints in Instructions over concepts in Library $\mathcal{L}$ and learning additional concepts via SPG if necessary.}
    \label{fig:constraint-wrapper}
\end{figure}

\textbf{Goal Grounding:}
A complex  Natural Language (NL) instruction is passed to the Parser, which extracts two components: the relevant concepts and the associated constraints. If some of the concept(s) are not present in the concept library  $\mathcal{L}$, the system queries for a Human Demonstration, which is then processed via the SPG framework, and the abstracted \& generalised concept(s) are appended to $\mathcal{L}$, as in Figure:~\ref{fig:constraint-wrapper}.

Following this, an LLM identifies whether an explicit constraint resolution is required. If yes, the Language Parser isolates the constraint components from the NL Instruction. , and generates a program of constraints that can be solved independently. This bifurcation allows the system to resolve increasingly complex constraints without entangling them with the core structural learning. This is an improvement over directly relying on LLMs to simultaneously learn structures and resolve complex constraints, since LLMs often fail to reason about spatial grounding, visibility, occlusion, and physical stability (Appendix:~\ref{app:planning-details}). Our pipeline modularizes the constraint resolution stage by invoking neural solver modules, external constraint solvers, or LLMs, based on downstream tasks, thereby improving the pure-LLM baseline. We aim to improve the robustness of our pipeline by integrating advanced reasoning capabilities.


\label{sec:application_plan_synthesis}
\textbf{Plan Synthesis:} We qualitatively demonstrate that the concepts we have acquired can help us to perform goal conditioned planning. These goals may either be direct instances of concepts in $\mathcal{L}$, or constraint-satisfying goals obtained from sec.~\ref{sec:applications_goal_grounding} above. We perform a forward search using the abstract state and action representations to the inferred goal akin to ~\citep{liu2024learning, kalithasan2024learningrecoverplanexecution}  (detailed in Appendix~\ref{app:planning-details}).


\section{Evaluation Setup} \label{sec:eval_setup}
\textbf{Corpus.} A corpus is created using a simulated Robot Manipulator assembling spatial structures on a table-top, viewed by a visual-depth sensor. Demonstration data ($3$ demonstrations per structure, up to $20$ objects per scene) includes RGBD observations of the action sequences (picking and placing of blocks) resulting in the construction of the final assembly using varied block instances and types (cubes, dice, lego etc.).
\begin{figure}[h!]
    \centering
    \includegraphics[width=0.42\textwidth]{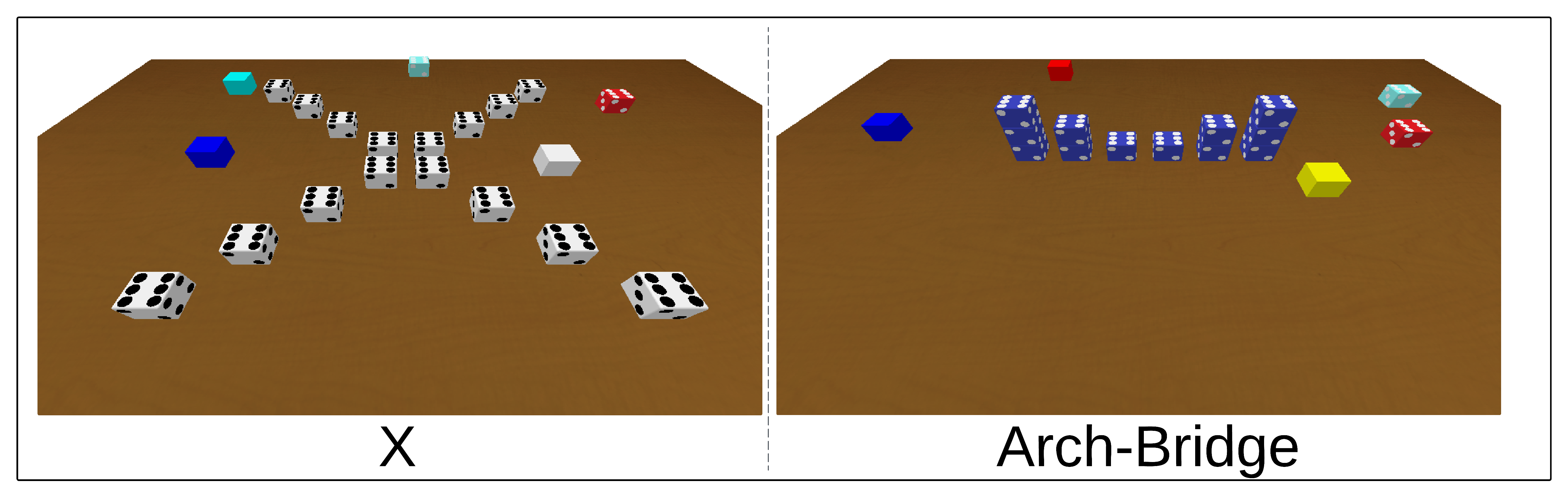}
    \caption{
    Illustrative examples of spatial structures from the corpus, showing inductive composition over simpler structures. 
    Details and visualizations in Appendix ~\ref{app:dataset}}
    \label{fig:example-structures}
    \vspace{-3mm}
\end{figure}
The scope of concepts and associated evaluation tasks are adapted from closely related works. The staircase and enclosure construction tasks are inspired from from~\cite{lpp}, adapted to 3D from the original 2D grid world setting. Structures such as boundaries involving repetitive use of columns and rows (w/o explicit joint fastening) are inspired by a robotic assembly data set~\citep{ramp}. Finally, the arch-bridge and x-shaped patterns are inspired from concept learning works as ~\cite{lake2015human}. A total of $15$ structures types are incorporated and are additionally modulated in size/spatial arrangement for generalization evaluation. 
%
Three evaluation data sets are formed each with \emph{simple} structures and \emph{complex} structures composed of simpler concepts (e.g., staircase consists of towers as substructure). \emph{Dataset I} and \emph{II} contain demonstrations constructing structures with $\mathtt{size}(.)   \in [3, 5]$, where $\mathtt{size}$ is defined in ~\ref{sec:representation}. \emph{Dataset II} reverses the linguistic labels used (e.g., the \emph{``tower"} in \emph{I} becomes \emph{``rewot"} in \emph{II}) to assess model reliance on pre-training knowledge in presence of new labels for concepts.  \emph{Dataset III} includes concepts of larger size than those in training to test generalization. 

\textbf{Baselines.} Four baselines are formed from two alternative approaches as follows. 

(1) \textit{Purely-Neural}: An end-to-end neural model inspired by StructDiffusion \citep{struct-diffusion} that treats structure construction as a rearrangement problem. We consider two variations: (1.1) Struct-Diff+Grounder(SD+G): End-to-end approach that assumes a perfect object selector/grounder which identifies the relevant set of objects to be moved. (1.2) Struct-Diff(SD): (1.1) without assuming a grounder. 

 
(2) \textit{Pre-trained models that directly output symbolic programs: }(2.1) LLMs for Scene-Graph Reasoning: This approach uses a Pre-trained Language Model (GPT-4) to generate Python programs from instructions which describe the given demonstration. Input is provided to the LLM via textual symbolic spatial relationships (e.g., left(a,b)) between objects in the demonstration. This baseline further assumes the absence of distractor objects in the scene. (2.2) Vision Language Model (GPT-4V): Similar to (2.1) but has the ability to take input demonstration as images. For learning the program of a new inductive concept, we give the demonstration to the VLM in the form of $\Lambda, (S_{1}..S_{g})$. Additional details on prompting method in Appendix~\ref{vlm-llm-baseline}. Experiments were also conducted with open-source LLMs such as CodeLlama (70Bq), Due to significantly poorer performances w.r.t. GPT-4, GPT-4 was retained as the primary LLM baseline. 

\textbf{Model variants.}
We implement three variants of the MCTS search to perform a grounded plan search over the action space $\mathcal{A}$: (i) MCTS+\emph{P}+\emph{L}: Our approach as described in section~\ref{subsec:mcts}, that uses the learnt concepts (\emph{L}) from $\mathcal{L}$ in subsequent searches, along with  pruning (\emph{P}) $\mathcal{A}_{p}$ using $\pi_{\text{neural}}$, (ii) MCTS-\emph{P}+\emph{L}: Our approach without neural pruning and (iii) MCTS+\emph{P}-\emph{L}: No access to library of concepts during continual learning. This method greedily selects the action from $\mathcal{A}_{p}$ as given by $\pi_{\text{neural}}$. For details refer App.~\ref{app:diff-plan-search-details}.

\textbf{Metrics.}  We adopt the following metrics to evaluate our models: (i)\textit{ Program Accuracy:} A binary score obtained through human evaluation. $1$ for constructing the structure fully, $0$ otherwise. (ii)\textit{ Target Construction IoU:} Intersection over Union (2D-IoU) between bounding boxes. (iii)\textit{ Target Construction Loss:} Mean Squared Error (MSE) loss over the bounding boxes + depth of the center. 


\section{Results}

Our experiments evaluate the following questions. \textbf{Q1:} How does our model perform when compared to baselines in terms of concept learning and execution ability (In-Distribution)? \textbf{Q2:} How does our model generalize to concept instances not seen (larger) during training (Out-of-Distribution)?  \textbf{Q3:} How robust and efficient is our concept learning pipeline? \textbf{Q4:} How can the acquired concepts be used in for embodied instruction following tasks?
\subsection*{\textbf{Q1}: Concept Learning Accuracy} 
We compare the program accuracy (Table~\ref{tab:prog_acc}) and the IoU/MSE values (Table~\ref{table:in-dist}) of the final states attained by SPG and the baselines w.r.t. the gold states in the in-distribution setting and find that SPG significantly outperforms other approaches. Values for Purely neural approaches are marked NA because Neural Outputs are not physically grounded. We make the following observations: (i) For \textit{complex} compositional structures, the accuracy of the pre-trained models is poor (zero), indicating their inability to reason over the numerous and complex spatial relations present in these structures. (ii) While program inference via the LLM is better than the VLM for learning \textit{simple} structures, it is worse for \textit{complex} structures. This indicates the inherent weakness of the textual descriptions of complex spatial relations present in complex structures. (iii) While the data-intensive purely neural approaches perform much better on \textit{complex} structures when compared to the pre-trained foundation models, they are still weaker than SPG.  
\begin{table}[h!]
\vspace{-2mm}
\caption{Program Accuracy}
\label{tab:prog_acc}
\centering
 \resizebox{0.25\textwidth}{!}{
    \begin{tabular}{l c c}
        \toprule
        \textbf{Model} & \textbf{Simple} & \textbf{Complex} \\
        \midrule
        SPG(Ours) & \textbf{1.00} & \textbf{0.83} \\
        GPT-4V & 0.33 & 0.00 \\
        GPT-4 & 0.78 & 0.00 \\
        SD+G & NA & NA \\
        SD & NA & NA \\
        \bottomrule
    \end{tabular}
}
\vspace{-4mm}
\end{table}
\begin{table}[h!]
\centering
\caption{In-distribution Performance (Mean $\pm$ Std-error)}
\vspace{-2mm}
\label{table:in-dist}
\centering
 \resizebox{0.49\textwidth}{!}{
    \begin{tabular}{l c c c c}
        \toprule
        \multirow{2}{*}{\textbf{Model}} & \multicolumn{2}{c}{\textbf{Simple}} & \multicolumn{2}{c}{\textbf{Complex}} \\
        \cmidrule(lr){2-3} \cmidrule(lr){4-5}
        & \textbf{IoU} & \textbf{MSE (1e-3)} & \textbf{IoU} & \textbf{MSE (1e-3)} \\
        \midrule
        SPG(Ours) & \textbf{0.96} $\pm$ 0.00 & \textbf{0.01} $\pm$ 0.00 & \textbf{0.85} $\pm$ 0.02 & \textbf{2.06} $\pm$ 1.02 \\
        GPT-4V & 0.75 $\pm$ 0.01 & 4.33 $\pm$ 0.41 & 0.50 $\pm$ 0.02 & 7.29 $\pm$ 1.10 \\
        GPT-4 & 0.89 $\pm$ 0.01 & 1.36 $\pm$ 0.26 & 0.28 $\pm$ 0.02 & 13.5 $\pm$ 1.65 \\
        SD+G & 0.74 $\pm$ 0.01 & 1.42 $\pm$ 0.29 & 0.61 $\pm$ 0.02 & 2.43 $\pm$ 0.48 \\
        SD & 0.49 $\pm$ 0.01 & 1.48 $\pm$ 0.24 & 0.46 $\pm$ 0.02 & 3.71 $\pm$ 1.53 \\
        \bottomrule
    \end{tabular}
}    

\end{table}
\vspace{-2mm}

\subsection*{\textbf{Q2: } Generalization Performance}
Table~\ref{table:out-dist}, compares the generalization performance on \emph{Dataset III} for models trained on \emph{Dataset I} (full table in Appendix, ~\ref{table:out-dist-std}). We see that SPG outperforms other approaches. We further consider the relative decrease (R.D.) in performance (2D-IoU) on going from the in-distribution to the out-of-distribution (OOD) setting. We make the following observations: (i) SPG suffers a R.D. of 7.27\% for simple and 5.74\% for complex structures. (ii) In contrast, the SD+G baseline shows a large R.D. of 63.25\% on simple structures and 74.72\% on complex structures; highlighting the inability of Purely Neural Models to generalize inductively. (iii) Pre-trained models also have a large R.D. in perf. for complex structures (GPT-4 : 53.87\% \& GPT4V : 41.64\%), which is attributed to their inability to generate the correct program that can generalize inductively to unseen data. 
%
%
\begin{table}[h!]
\caption{OOD Performance. \footnotesize{R.D\% is the relative decrease in IoU from Table~\ref{table:in-dist}. MSE is in 1e-3 units}}
\label{table:out-dist}
\centering
\resizebox{0.45\textwidth}{!}{ 
    \begin{tabular}{l c c c c c c}
        \toprule
        \multirow{2}{*}{\textbf{Model}} & \multicolumn{3}{c}{\textbf{Simple}} & \multicolumn{3}{c}{\textbf{Complex}} \\
        \cmidrule(lr){2-4} \cmidrule(lr){5-7}
        & \textbf{IoU} & \textbf{R.D\%} & \textbf{MSE} & \textbf{IoU} &  \textbf{R.D\%} & \textbf{MSE} \\
        \midrule
        SPG(Ours) & \textbf{0.89} & \textbf{7.27} & \textbf{0.43} & \textbf{0.80} & \textbf{5.74} & \textbf{1.49} \\
        GPT-4V    & 0.58 & 23.33 & 13.2 & 0.29& 41.64 & 10.9 \\
        GPT-4     & 0.78 & 12.61 & 5.51 & 0.13& 53.87 & 19.1 \\
        SD+G      & 0.27 & 63.25 & 6.21 & 0.15& 74.72 & 14.2 \\
        SD        & 0.24 & 51.84 & 6.86 & 0.15 & 67.67 & 11.6 \\
        \bottomrule
    \end{tabular}
}
\end{table}
\subsection*{\textbf{Q3: }Robustness and Efficiency Analysis}
\textbf{Reliance on pre-trained Knowledge vs. Demonstration.} 
Next, we evaluate the degree to which concept learning relies on prior knowledge vs. the action sequences observed in demonstrations. We compare pre-trained models against our approach by learning programs on \textit{Dataset II} (~\ref{sec:eval_setup}), which uses arbitrary names for concepts. This forces all models to rely on demonstration data because there is no real-world knowledge associated with the name of the concept, say  \textit{"rewot"} instead of \textit{"tower"}. Table ~\ref{table:in-dis-nr} indicates the corresponding performances, with our approach outperforming others. For the IoU/MSE values along with standard errors refer to app. Table ~\ref{table:in-dis-nr-std}. The R.D. in performance (program accuracy w.r.t. Table ~\ref{tab:prog_acc}) for \textit{simple} structures  for our approach (12\%) is lower than GPT-4 (14\%) \& much lower than GPT-4V (30\%). The poorer generalization of pre-trained models can be attributed to their over-reliance on prior knowledge and failure to effectively incorporate the data from demonstrations. In contrast, SPG better captures the semantics of a novel concept, especially ones whose knowledge may not be available for the LLMs/VLMs at training time.

\begin{table}[H]
\caption{Perf. on \textit{Dataset II} with Reversed Names. \footnotesize{Acc. is Prog. Accuracy, MSE in 1e-3 units}}
\label{table:in-dis-nr}
\vspace*{-0.3cm}
\centering
\resizebox{0.48\textwidth}{!}{ 
\begin{tabular}{l c c c c c c}
    \toprule
    \multirow{2}{*}{\textbf{Model}} & \multicolumn{3}{c}{\textbf{Simple}} & \multicolumn{3}{c}{\textbf{Complex}} \\
    \cmidrule(lr){2-4} \cmidrule(lr){5-7}
     & \textbf{Acc.} & \textbf{IoU} & \textbf{MSE} & \textbf{Acc.} & \textbf{IoU} & \textbf{MSE} \\
    \midrule
    SPG(Ours) & \textbf{0.88} & \textbf{0.86} & \textbf{1.74} & \textbf{0.78} & \textbf{0.78} & \textbf{3.93} \\
    GPT-4V & 0.23 & 0.71 & 3.92 & 0.00 & 0.09 & 21.29 \\
    GPT-4  & 0.67 & 0.78 & 3.16 & 0.00 & 0.00 & 22.73 \\
    \bottomrule
\end{tabular}
}

\vspace*{-0.3cm}
\end{table}

\begin{figure}[H] 
    \centering
    \includegraphics[width=0.40\textwidth, height=0.45\linewidth]{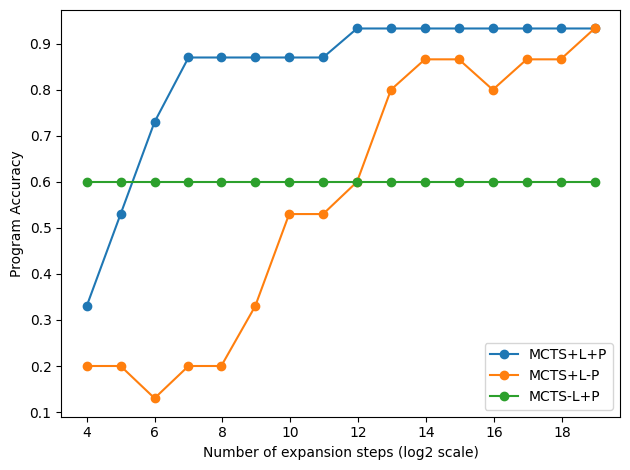}
    \caption{
    \footnotesize{\textbf{MCTS Variants.} Num. of expansion steps in search (log scale) (X-axis) vs Program accuracy (Y-axis).}}
    \label{fig:diff-plan-search}
        \vspace{-3mm}
\end{figure}

\textbf{MCTS Variants for Concept Learning.}
Figure ~\ref{fig:diff-plan-search} compares the program accuracy for the three methods of plan search. For the MCTS-\emph{L}+\emph{P} method, the program accuracy is expected to be independent of expansion steps as it greedily chooses the action for which $\pi_{\text{neural}}$ gives the highest probability. For the MCTS+\emph{L} based methods the accuracy increases beyond 0.6 with time, which demonstrates that having a composable library of concepts allows us to learn a much richer class of inductive concepts. MCTS+\emph{P}+\emph{L} saturates to a program accuracy of 0.933 in just 4000 expansion steps compared to  MCTS+\emph{L}-\emph{P} taking 512000 expansion steps, which demonstrates a significant increase in learning efficiency via use of the neural pruner. For very low number of expansions steps (<40) accuracy of MCTS+\emph{L} based methods is lower than MCTS-\emph{L} as the former expends expansion steps on UCB exploration (instead of greedy actions).

\textbf{Significance of MCTS in SPG.}  
To assess the necessity and importance of MCTS, we carry out an ablation study where we replace it with an LLM planner during the planning stage, referred to as SPG-M+LMP. In the "plan" stage of our pipeline, GPT-4V is prompted to output a plan given the concept library and RGB keyframes from the demonstration. Our experiment shows that GPT-4V struggles to generate correct plans, particularly for complex structures like pyramids, arch\_bridge and boundaries, resulting in significantly lower performance than SPG (Table~\ref{fig:ablation_studies}). Additionally, some plans generated by GPT-4V are not physically grounded, leading to errors in both the planning and generalization stages, which compounds the inaccuracies.  This demonstrates that combining symbolic search with LLMs offers a substantial advantage over using only LLMs.
\begin{table}
    \caption{Ablation studies. \footnotesize{Ablations with SPG-M+LMP and GPT-4V+VR. MSE values are in 1e-3.}}
    \centering
     \vspace{-3mm}
    \resizebox{0.48\textwidth}{!}{ 
        \begin{tabular}{l c c c c c c}
            \toprule
            \multirow{2}{*}{\textbf{Model}} & \multicolumn{3}{c}{\textbf{Simple}} & \multicolumn{3}{c}{\textbf{Complex}} \\
            \cmidrule(lr){2-4} \cmidrule(lr){5-7}
            & \textbf{Acc.} & \textbf{IoU} &  \textbf{MSE} & \textbf{Acc.} & \textbf{IoU} &  \textbf{MSE} \\
            \midrule
            SPG(Ours)  & \textbf{1.0} &	\textbf{0.96} &	\textbf{0.01} & \textbf{0.83} & \textbf{0.85} & \textbf{2.06} \\
            SPG-M+LMP  & 0.55 &	0.68 & 11.1	& 0.16 & 0.19 & 20.0 \\
            GPT-4V+VRF  & 0.66 & 0.75 &  6.8 & 0.16 & 0.46 & 12.0 \\
            \bottomrule
        \end{tabular}
    }
    \label{fig:ablation_studies}
\end{table}
\textbf{Ablating with Pre-trained Models + Visual Reward Filter}
In line with program synthesis techniques using LLMs \citep{Li_2022,chen2021evaluatinglargelanguagemodels}, we sample five programs from GPT-4V and rank them according to the visual reward obtained from their execution. Furthermore, we provide the ground-truth programs of the concepts that are needed to learn the given new concept, thus employing a form of teacher forcing in program generation. Even with these measures, it performs significantly worse than SPG, see Table~\ref{fig:ablation_studies} (GPT-4V+VRF). While this performance is better than that of GPT-4V, it still unable to generate correct programs, especially for complex structures.  
\subsection*{Q4: Incorporating Constraints into learnt concepts }
\label{sec:applications}
\begin{figure}[h]
    \centering
    \includegraphics[width=0.2\textwidth]{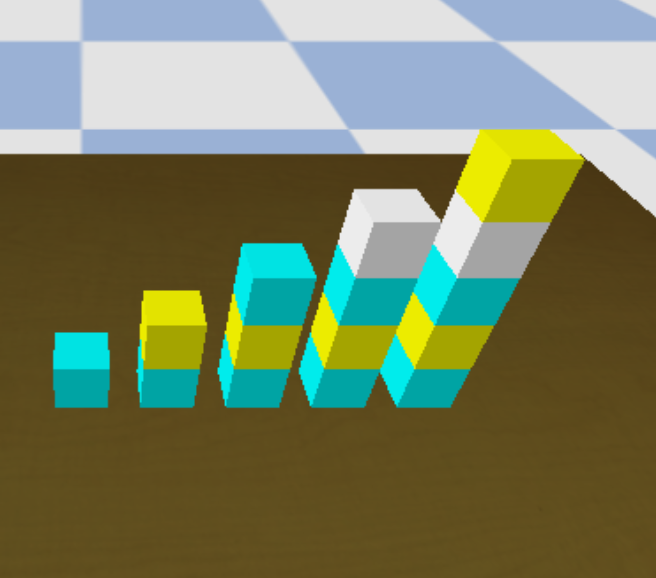}
    \caption{\footnotesize{Constructed staircase obeying all constraints in $\mathcal{\lambda}_1$}, which are extracted via an LLM and solved via Z3 CSP-solver.}
    \label{fig:staircase-sim}
\end{figure}

\begin{figure}[h!]
 \vspace{-3mm}
    \centering
    \includegraphics[width=0.48\textwidth, trim={0cm 13.5cm 0 2.8cm},clip]{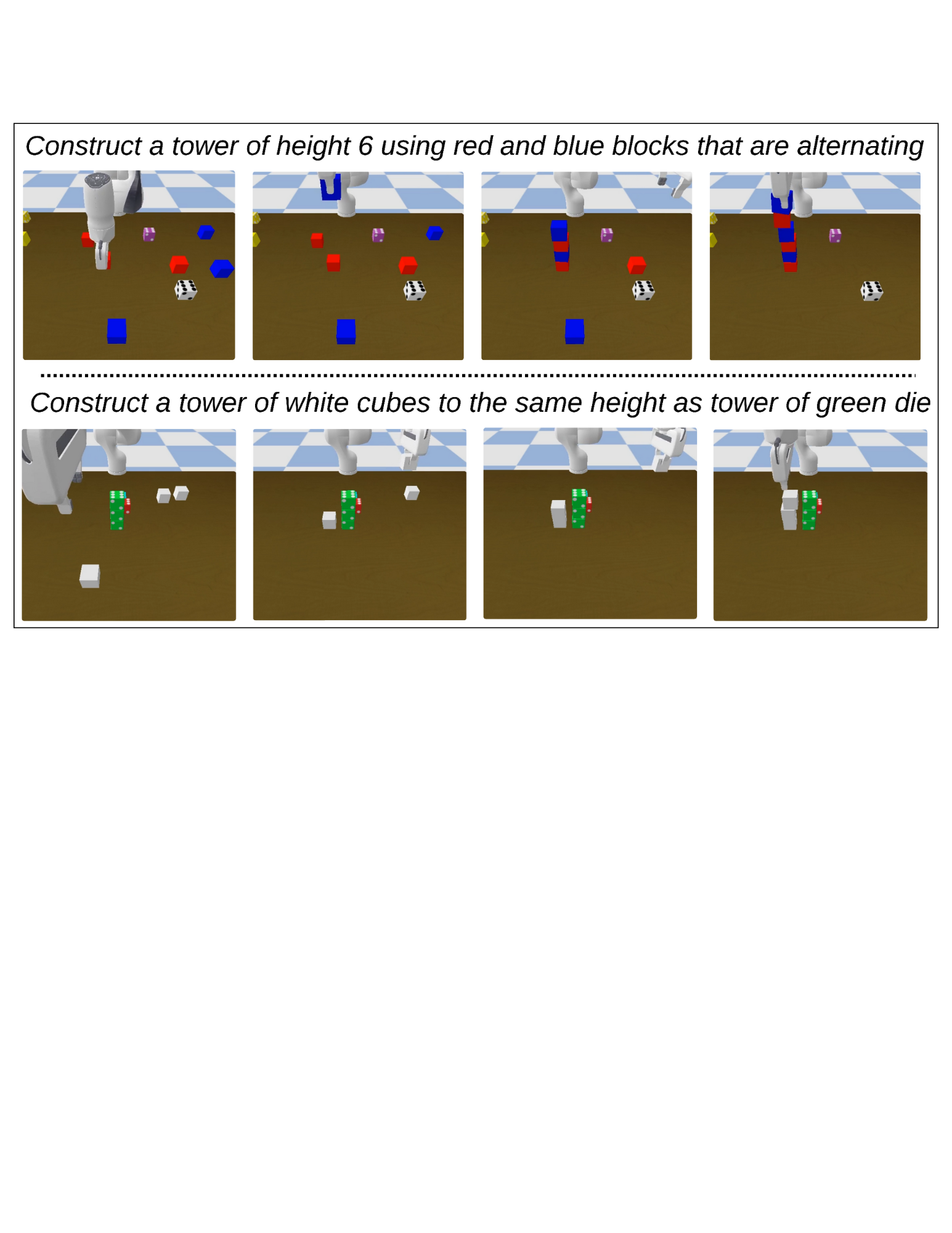}
    \caption{
     \footnotesize{Using an LLM to solve the constraints in a novel task and generate executable code, given the concept definitions.}}
    \label{fig:qual_results_1}
\end{figure}

\textbf{Instructions with Discrete Constraints:}Consider the instruction $\mathcal{\lambda}_1$: \textit{``Construct a staircase of size 5 such that all blocks have the same color as the block to their left. No block should have the same color as the block on top of it.''} This example highlights the interpretive \& compositional strength of our pipeline. We model the constraints as a CSP problem \& use the Z3 solver \citep{Moura2008Z3Solver} to resolve them. The resulting grounded program for $\mathcal{\lambda}_1$ is executed via the planner, producing the desired 3D structure. Fig~\ref{fig:staircase-sim} shows the executed output in the simulator, satisfying both the structural concept and the imposed constraints.
%
%
For simpler instructions such as  $\mathcal{\lambda}_2$:
 \textit{''Construct a tower of green die having the same height as the existing tower of white die.''} and $\mathcal{\lambda}_3:$\textit{''Construct a tower of total 6 blocks using alternating blue and red blocks.''}, the LLM (GPT-4) is able to understand the structural requirements (through the library $\mathcal{L}$) as well as the (simpler) constraints. Therefore, the pipeline (with access to \texttt{tower} through $\mathcal{L}$ or via SPG framework), generates Python code in terms of these functions, which on execution generates the required action sequence as shown in Fig.~\ref{fig:qual_results_1}. (also see app. ~\ref{details-hri}). 

\begin{figure}[h!]
    \vspace{-3mm}
    \centering
    \includegraphics[width=0.48\textwidth, trim={0cm 12.0cm 0 3.8cm},clip]{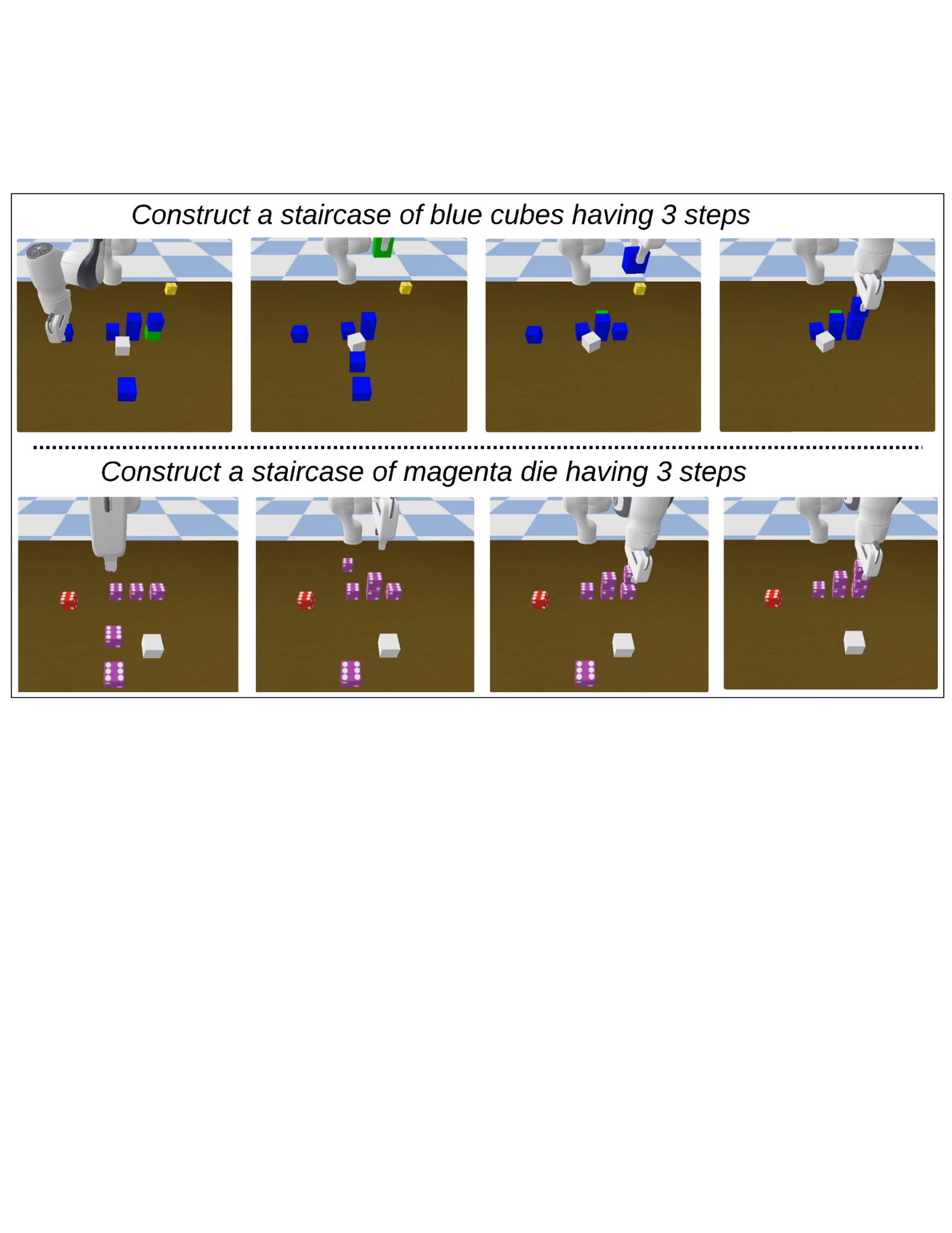}
    \caption{
     \footnotesize{Integrating a neuro-symbolic planner over the concepts. Top: The planner is able to optimally replace the green cube from the adversarial initial state by unstacking and re-stacking the faulty tower. Bottom: The planner is able to complete a staircase from an initially constructed row by layering rows upon rows, a method of construction it has not seen while learning staircase.}}
    \label{fig:qual_results_2}
     \vspace{-3mm}
\end{figure}

\textbf{Efficient Plan Synthesis for Grounded Goals:} As detailed in sec.~\ref{sec:application_plan_synthesis} and App.~\ref{app:planning-details}
, we demonstrate the ability to perform goal-conditioned planning using the acquired concepts. Fig~\ref{fig:qual_results_2} demonstrates the results of our approach for the tasks of constructing a staircase beginning from adversarial and assistive initial states.

\section{Conclusion}
This paper introduces a novel approach for learning inductive representation of grounded spatial concepts as neuro-symbolic \emph{programs} via language-guided demonstrations. 
Our approach factors program learning as: {\it Sketch:} generating the high-level program signature via an LLM, {\it Plan:} searching for a grounded plan that maximises the total discounted reward with the respect to the demonstration, and {\it Generalize:} abstracting the grounded plan into an inductively generalize-able abstract plan via an LLM. Continual learning is achieved via learning of modular programs by giving preference to shorter programs through composition of learnt ones.
Extensive evaluation demonstrates accurate program learning and stronger generalization in relation to purely LLM based as well as purely neural baselines. Grounding of learned concepts in visual data facilitates reasoning and planning for embodied instruction following. Limitations include reliance on perfect demonstrations, assumption of full observability of all objects and experiments confined to simulation. Incorporating noisy demonstrations, reasoning with beliefs and interleaving planning and execution remains part of future work.


\bibliography{references}
\bibliographystyle{icml2025}

\newpage
\appendix
\onecolumn

\appendix
\onecolumn


\section{Additional Details on Technical Approach}

\begin{figure}[hbt!]
    \centering
    \includegraphics[width = \textwidth]{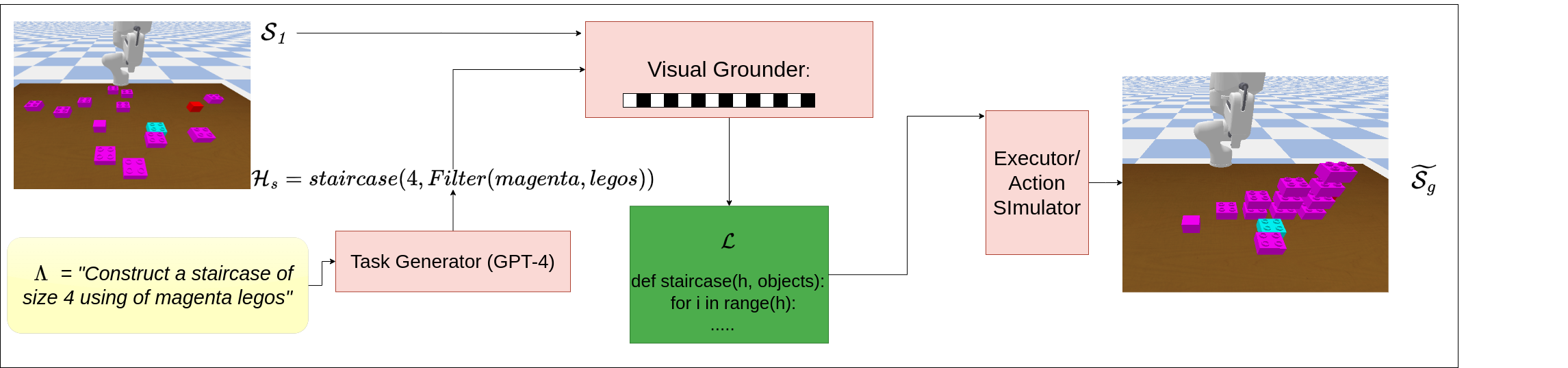}
    \caption{\textbf{SPG: Inference} First the library of concept $\mathcal{L}$ is loaded with the corresponding set of learnt programs. Then the given instruction is converted into task-sketch $H_{s}$, which is grounded in the initial scene. The required program is fetched from the library, and the grounded task-sketch is executed based on the semantics of the learnt program.} 
    \label{fig:inference}
\end{figure}

Figure ~\ref{fig:inference} illustrates the pipeline for online inference to realize to realize construction of novel structures. 

\subsection{Symbolic Constructs and their Semantics used in Programs} 

\label{app:symbols-and-semantics}
Table \ref{tab:operators} defines the types of the signature and semantics of all the operators. Table \ref{tab:types}, includes the type definition of various symbols. Standard Python constructs such as for loops, if else $\cdots$) as assumed in addition to the constructs defined here. 

\begin{table}[h!]
    \centering
      \caption{\textbf{Symbols and Semantics} Signature and semantics for the primitive concepts and operations that are used in the construction of the programs used to express inductive spatial concepts.}
    \begin{tabular}{|p{4cm}|p{5cm}|p{4cm}|}
    \hline
        \textbf{Function} &  \textbf{Signature} & \textbf{Semantics} \\
        \hline
        filter &  (VisualConcept, ObjSet) $\rightarrow$ ObjSet  &
         Returns the objects that contain the VisualConcept\\
         \hline
         move\_head & (Head, Dir) $\rightarrow$   Head & Moves the head to the given direction (May or may not take input/return the head, based on a flag)\\
         \hline
         assign\_head a.k.a move\_head(overloaded) & (Head, ObjIdx) $\rightarrow$   Head & Given the index/one-hot representation for an object, it moves the head to the position corresponding to that object\\
         \hline
         keep\_at\_head & (ObjSet, Head) $\rightarrow$ None &  Keeps the argmax of ObjSet at the head\\
         \hline
         reset\_head & None $\rightarrow$ Head &  Sets the head to the top position of stack and pops this position from the stack as well\\
         \hline
         store\_head & Head $\rightarrow$ None &  Pushes the current position of head into the stack\\
         \hline
         
    \end{tabular}
    \label{tab:operators}
\end{table}

\begin{table}
    \centering    
    \caption{\textbf{Symbolic representation.} The table lists the type definitions used in the implementation of SPG programs. }
    \begin{tabular}{|p{3cm}|p{3cm}|p{5cm}|}
    \hline
        \textbf{Defined Types} &  \textbf{Python Type}  & \textbf{Usage} \\
        \hline
        IntArg & int &  Argument for the structures that defines the size (height, length etc) \\
        \hline
        Obj & torch.Tensor & One-hot vector whose non-zero index represents the selected objects \\
        \hline
        ObjSet & torch.tensor & Probability mask over the selected objects \\
        \hline
        Dir & string & Primitive directions like left, right, front, top, etc \\
        \hline
        ConceptName & string & name of the visual, action or inductive concept \\
        \hline
        Head & torch.Tensor & Bounding box with depth. 3D cuboidal space. \\
        \hline
    \end{tabular}
    \label{tab:types}
\end{table}

\subsection{Curriculum Learning} 
\label{curriculum}
\begin{figure}[hbt!]
    \centering
    \includegraphics[width=\textwidth]{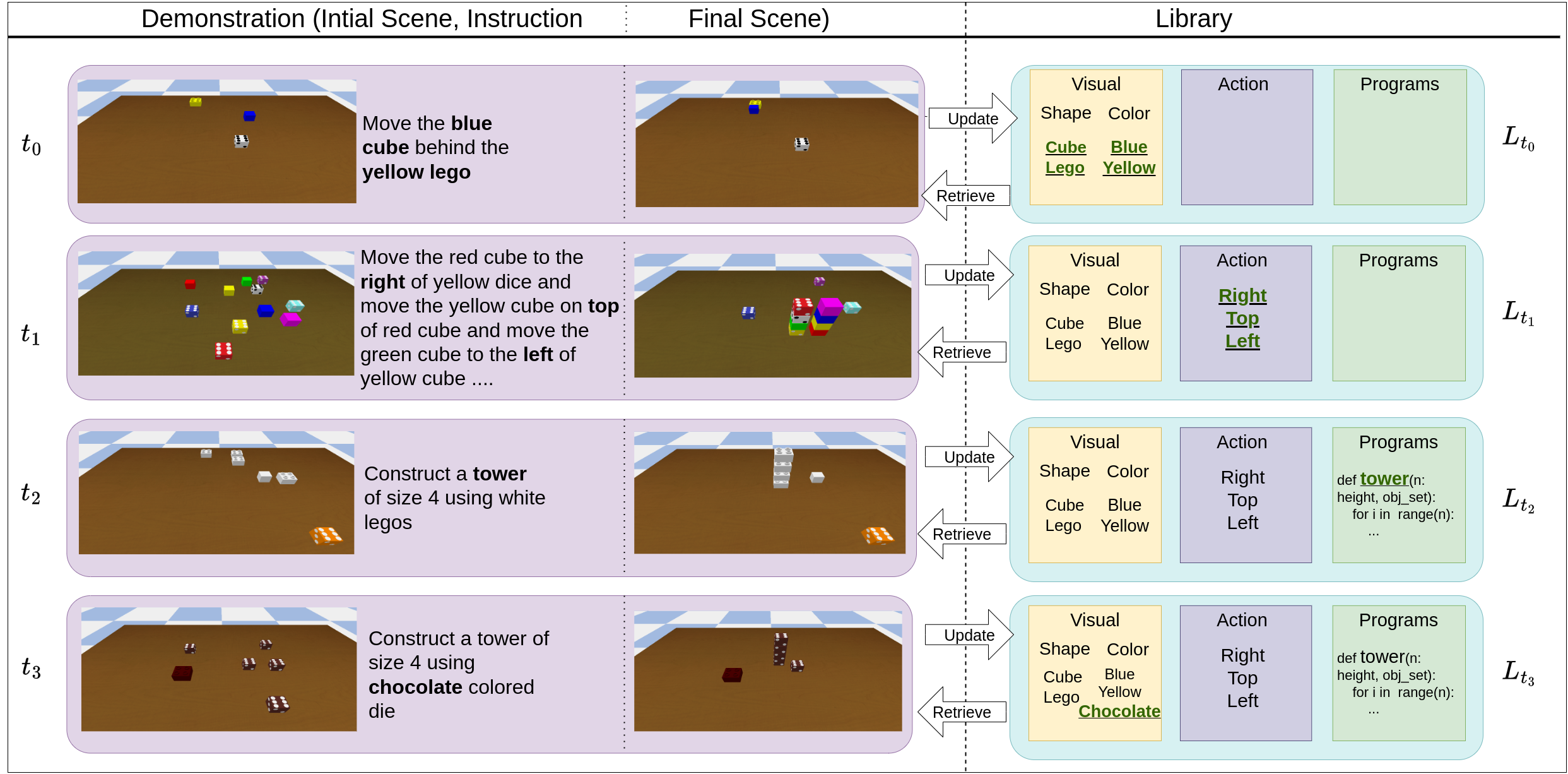}
    \caption{
   \textbf{Continual learning through curriculum:} Using simple pick and place demonstrations we learn visual attributes such as \emph{blue cube, yellow lego} ($t_{0}$). Using long range instructions which are sequentially concatenated descriptions of pick and place tasks we train our action concepts such as \emph{left, right, top} ($t_{1}$). After pre-training the agent can perform continual learning of concepts such as learning generalized representation for \emph{tower} ($t_{2}$). Because of disentangled representation of neural and symbolic concepts, interspersed learning of new visual attributes such as \emph{chocolate color} are also possible through few demonstrations of structure creation($t_{3}$).}
    \label{fig:training-details}
\end{figure}

 We follow a curriculum approach where the visual concepts are trained first from simpler linguistically-described demonstrations ($t_{0}$ in figure ~\ref{fig:training-details}). This is followed by learning of action concepts through sequentially composed pick and place tasks. Its essential to use such long range sequential instructions in order to ensure that the semantics of action concepts are learnt for placement of objects at a height much above the tabletop ($t_{1}$ in figure ~\ref{fig:training-details}). After the pre-training phase, the agent can continually learn new inductive concepts and visual attributes. ($t_{2}, t_{3}$ in figure ~\ref{fig:training-details}) 
\subsection{Details for Plan-Search and Generalization}
\label{search-gen-details}

\begin{figure}[h]
    \centering
    \includegraphics[width=0.5\linewidth, height = 12cm]{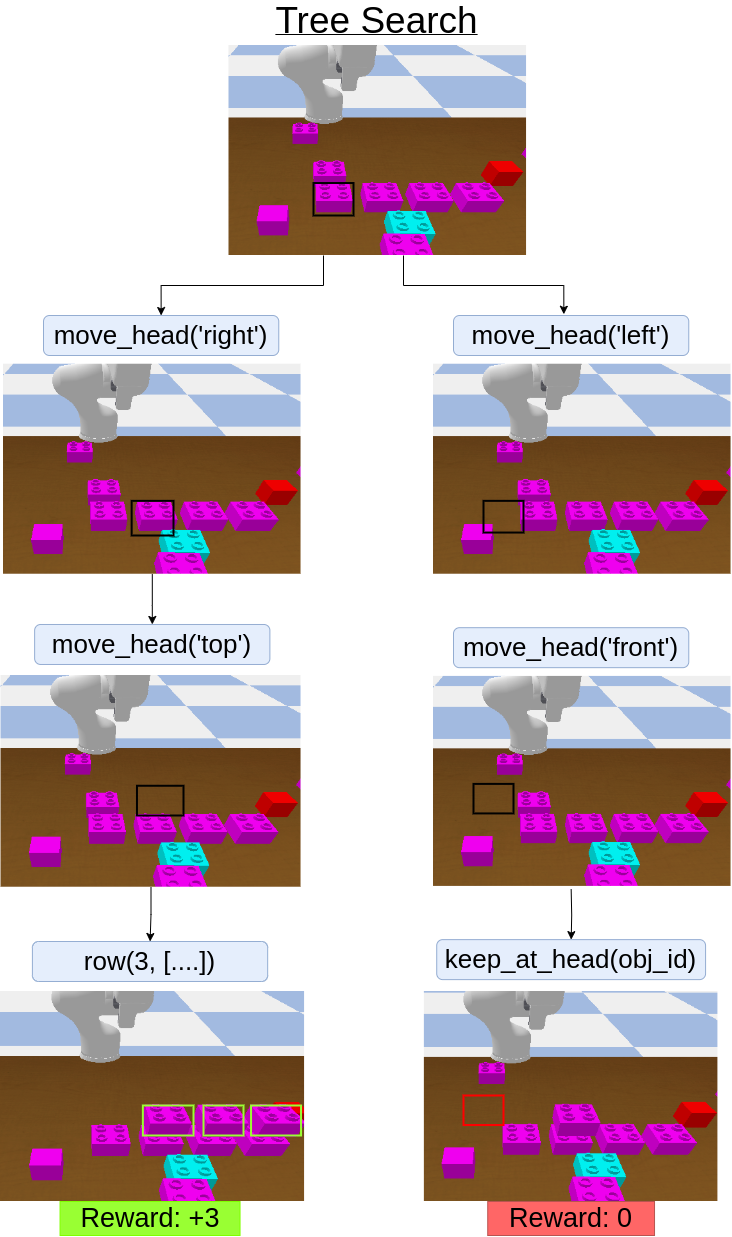}
    \caption{A sample MCTS search tree outlining the states explored and the calculation of reward.}
    \label{fig:sample_mcts_tree}
\end{figure}

\textbf{Modifications to the Simulation and Reward Back propagation Steps}:
Next, we outline the modifications in the simulation and the reward back propagation steps of the standard MCTS algorithm for our setting. During program search we assume access of intermediate scenes in the demonstration. This allows us to provide intermediate rewards that can guide the search well. We observed that making the following changes in simulation and back propagation step increased the efficiency of our search procedure. Fig \ref{fig:sample_mcts_tree} illustrates the possible states explored by MCTS and the reward calculation. 
\begin{itemize}
    \item Simulation: Rather than performing Monte Carlo simulations at each newly expanded leaf node (to estimate its value) we completely avoid these simulation steps. This was motivated by the fact that our reward is not completely sparse and the intermediate IoU rewards for each object we place allow us to guide the search effectively.
    \item Back propagation: We perform off policy Q-learning updates during back propagation similar to one indicated by \cite{complexmctsbackup} :
    \begin{equation}
        V(s_{t}) = max_{a \in \mathcal{A}}Q(s_{t}, a).
    \end{equation}
    \begin{equation}
        \tau(s_{t}, a) = s_{t+1}
    \end{equation}
    \begin{equation}
        Q(s_{t}, a) = r_{t} + \gamma V(s_{t+1})
    \end{equation}
\end{itemize}

\textbf{Improving Modularity and Scalability of MCTS procedure:} We present additional details on the MCTS procedure for searching for a plan conditioned on a program signature and guided by the demonstration. 

In order to incorporate the objective of searching for physically realizeable plans and facilitating generation via re-use of concepts, the following conceptual changes are incorporated in the standard MCTS procedure~\cite{sutton2018reinforcement}.  

\textbf{Modularity (MCTS + \emph{L})}: We want to allow learning of novel inductive concepts in terms of existing ones. This would ensure that the plan $H_P^*$ corresponding to a given demonstration is concise and can be easily generalized to the $H_G^*$. ~\ref{modular-plans} in appendix give example of two plans for the structure \emph{Pyramid} one which is modular and can be successfully generalized by GPT-4, other for which GPT-4 fails in generalization due to lack of modularity. This can be seen as a form of regularization in terms of the length of concept description, by making the prior $P(H) \propto |H|^{-\alpha}$ (where $\alpha$ > 0) in equation (2)
\begin{equation}
    H^* = \arg \min_{H \in \mathcal{H}} \left[ \mathtt{Loss}(\{S_{1}..S_{g}\}, H(\Lambda)) + \alpha \log |H| \right]
\end{equation}
    In order to allow modular learning of programs, for every inductive concept already stored in the library we define corresponding action instantiations which can be a potential candidate actions during our search. As an example, for the concept \texttt{Tower} we have one of the action instantiation as \texttt{Make\_Tower(3, objects)}  which would be the action of constructing desired tower. This can be visualized as a compound/macro-action which is composed of primitive actions \texttt{keep\_at\_head(objects), move\_head(`top')}. We define $\mathcal{A}_{c}$ as the space of such compound actions, and $\mathcal{A}_{p}$ as the space of primitive action/function consisting \texttt{reset\_head(), move\_head(direction), keep\_at\_head(objects), store\_head()}. Realization of equation (6) (increased preference of macro-actions over primitive ones) is done through discounted IoU rewards during our search (\texttt{Make\_Tower(3)} would have a reward of 1+1+1, as compared to 1 + $\gamma^2$(1) + $\gamma^4$(1) for a sequence of 3 (\texttt{keep\_at\_head(objects), move\_head('top')}0)).
    We refer to MCTS using concept instantiations from $\mathcal{L}$ as macro actions in search as MCTS+\emph{L}.
    
\textbf{Scalablility (MCTS+\emph{P})}: as more concepts are added to the library $\mathcal{L}$ the action space of our search $\mathcal{A}$  (specifically $\mathcal{A}_{c}$, the space of compound concepts) increases, therefore we want to prune the search space effectively. For this during the pre-training phase we train a reactive policy $\pi_{neural}$ which given the current state, $\tilde{s_{t}}$ and the next expected state ${s_{t+1}}$ (part of the demonstration) would output one of the primitive action, $a_{t}^* \in \mathcal{A}_{p}$ where $\mathcal{A}_{p}$ is the primitive action space, i.e. $a_{t}^* =  \pi_{neural}(a_{t} | \tilde{s_{t}}, {s_{t+1}}), \; \; a_{t} \in \mathcal{A}_{p}$
Note that while expanding our search tree we only search among the space of compound actions $\mathcal{A}_{c}$ and the action $a_{t}^*$, thereby reducing the branching factor of search from $|\mathcal{A}_{c} \cup \mathcal{A}_{p}|$ to $|\mathcal{A}_{c}| + 1$. We refer
to MCTS using neural pruning as MCTS+\emph{P}.
Therefore our MCTS algorithm is modular through hierarchical composition of learnt concepts and efficient through pruning of action space and is referred to as MCTS+\emph{L}+\emph{P}.

\textbf{Generalization}: GIven multiple equal length plans for a given demonstration, we seek to recover a plan one that can be easily generalized by the LLM. ~\ref{lst:correct_gen} shows a plan which could be correctly abstracted out into a generic program by GPT-4. Whereas ~\ref{lst:incorrect_gen} shows another plan with similar semantics, for which GPT-4 is unable to correctly find the generalized program (Note that row or column of size 1 is equivalent to keep\_at\_head). We tackle this problem in the following manner. 
\begin{enumerate}
    \item Rather than getting a single plan from the plan search we get the top k plans $\{H_{P, i}\}_{i=1}^{i=k}$. In order to get these top k plans we expand the complete tree (based on UCB criteria) starting from the root node corresponding to the initial state, till a predefined budget of expansions. Then we select the top k paths(potential plans) from the root node to all the leaf nodes (where the top k ones are those that give the highest accumulated IoU reward with respect to the given demonstration).
    \item Later we abstract out each of these k plans into corresponding generalized programs, $\{H_{G, i}\}_{i=1}^{i=k}$ using GPT-4. We again run each of these programs on the given demonstration and then choose the one which gives the highest IoU reward (resolving ties based on predefined order).
    Note that some program $H_{G, i}$ upon execution may result in a plan $\tilde{H}_{P, i}$ different from the original plan ${H}_{P, i}$ using which it was generalized. This can be attributed to potential errors in GPT-4s program generalization process.
\end{enumerate} 

\subsection{Additional Details: Learning with Increasing Number of Demonstrations}
\label{app:learning-from-multiple-dem}
Given \emph{k} demonstrations for a novel inductive concept, we independently find k task sketch $\{H_{S, i}^*\}_{i=1}^{i=k}$ and grounded plans $\{H_{P, i}^*\}_{i=1}^{i=k}$. During the generalization phase we give these k pair of task-sketch and corresponding plans to GPT-4 and ask into infer a single abstraction over them. ~\ref{gen-mul-plan} in appendix gives a concrete example. Equation for generalize step (getting $H_{G}^*$ from $H_{P}^*, H_{S}^*$) can be modified as follows. 
\begin{equation}
    H_G^* \leftarrow Generalize(H_G \, | \, {\{H_{P, i}^*, \, H_{S, i}^*\}}_{i=1}^{i=k}; \,{\theta_G}), \; \; H_G \in \mathcal{H}^\mathcal{G}
\end{equation}

\subsection{Detailed Experimental Methodology}
\label{sec:exp_methodology}
\textbf{Input to the Method:} The input consists of a language instruction and a human demonstration represented as a sequence of RGBD keyframes.

\textbf{Output/Aim of the Model:} The goal is to learn a representation of the unknown concept in the instruction, assuming there is only one unknown concept. If the unknown concept is inductive (e.g., "tower"), the model learns a program definition $def tower()$ and stores it in the program library. If the unknown concept is a primitive concept, a concept embedding is learned through backpropagation.

\textbf{Evaluation:} The learned model is evaluated based on the correctness of the program representation for inductive concepts and the correctness of object placements, measured through the Intersection over Union (IoU) metric (see Metrics, line 262).
d. Examples: Suppose the current library contains the concepts {“red”, “tower”}. Given the instruction “construct a staircase of height 3 using red blocks,” the process is as follows:

\textbf{Parsing:} The instruction is parsed into a sketch: Staircase(height=3, objects=filter(red, blocks)).

\textbf{Grounding:} The “objects” parameter is grounded using the visual grounder, which identifies the indices of the red blocks, e.g., [1, 2, 3, 4, 5, 6]. That is, filter(red, blocks) = [1,2,3,4,5,6].

\textbf{Planning:} The planning step uses the demonstrations (sequence of keyframes) to identify the sequence of actions that best explains the demonstrations. In this case, the plan might be: Tower(height=1, objects=[1,2,3,4,5,6]), move\_head(right), Tower(height=2, objects=[2,3,4,5,6]), move\_head(right), Tower(height=3, objects=[4,5,6]

\textbf{Generalization:} The generalization step abstracts the plan obtained from three such demonstrations into a program. The resulting program would be:
\begin{lstlisting}[language=Python]
def staircase(height, objects):
    for i in range(height):
        tower(height=i, objects)
        move_head(right)
\end{lstlisting}

\textit{Note: } Whenever an object is placed, the objects list is modified in place, and the index of the placed object is removed.
primitive actions: The movement of any object is achieved by first determining the placement position by moving the head (an imaginary bounding box) in specific directions and then placing the object to be moved at the head. The head is implemented as a 3D bounding box defined by coordinates (x1, y1, x2, y2, d), where x1, y1, x2, and y2 are the 2D corners of the bounding box, and d is the depth at the center. The primitive action move\_head(direction) shifts the bounding box in the required direction. The action keep\_at\_head(object\_list) picks the first object in the list and places it at the center of the bounding box. Two other primitives, store\_head and reset\_head, are used to save the current position of the head, allowing the search to return to useful positions later if needed.

\section{Additional details regarding datasets}
\label{app:concept-info}
\begin{figure}[hbt!]
    \centering
    \includegraphics[width = \textwidth]{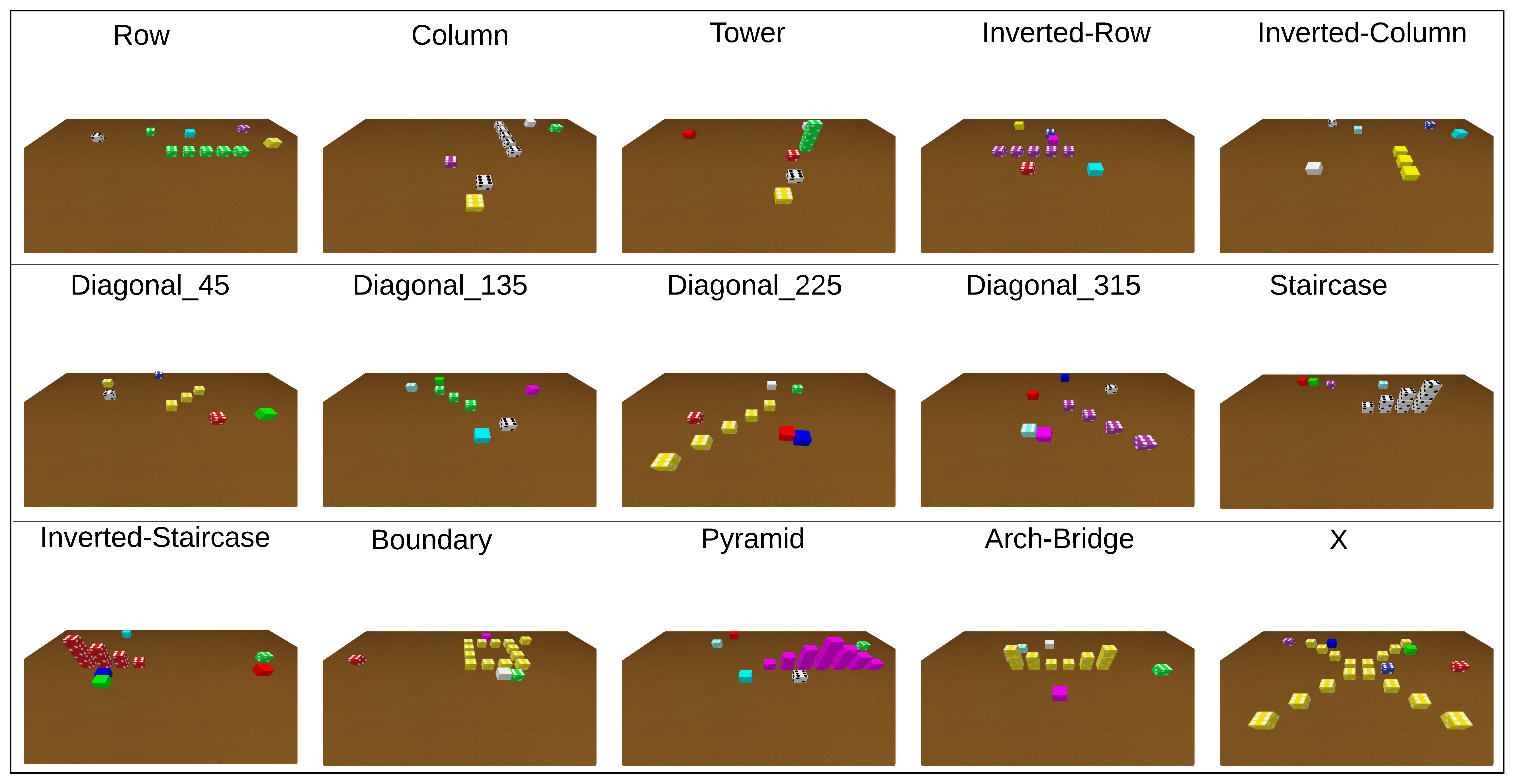}
    \caption{\textbf{Illustration of the inductive concepts.}} 
    \label{fig:struct-illus}
\end{figure}

Figure ~\ref{fig:struct-illus} demonstrates the kind of inductive concepts for which we want to learn generic (i.e instance agnostic) representations.

\begin{figure}[hbt!]
    \centering
    \includegraphics[width = 1\textwidth]{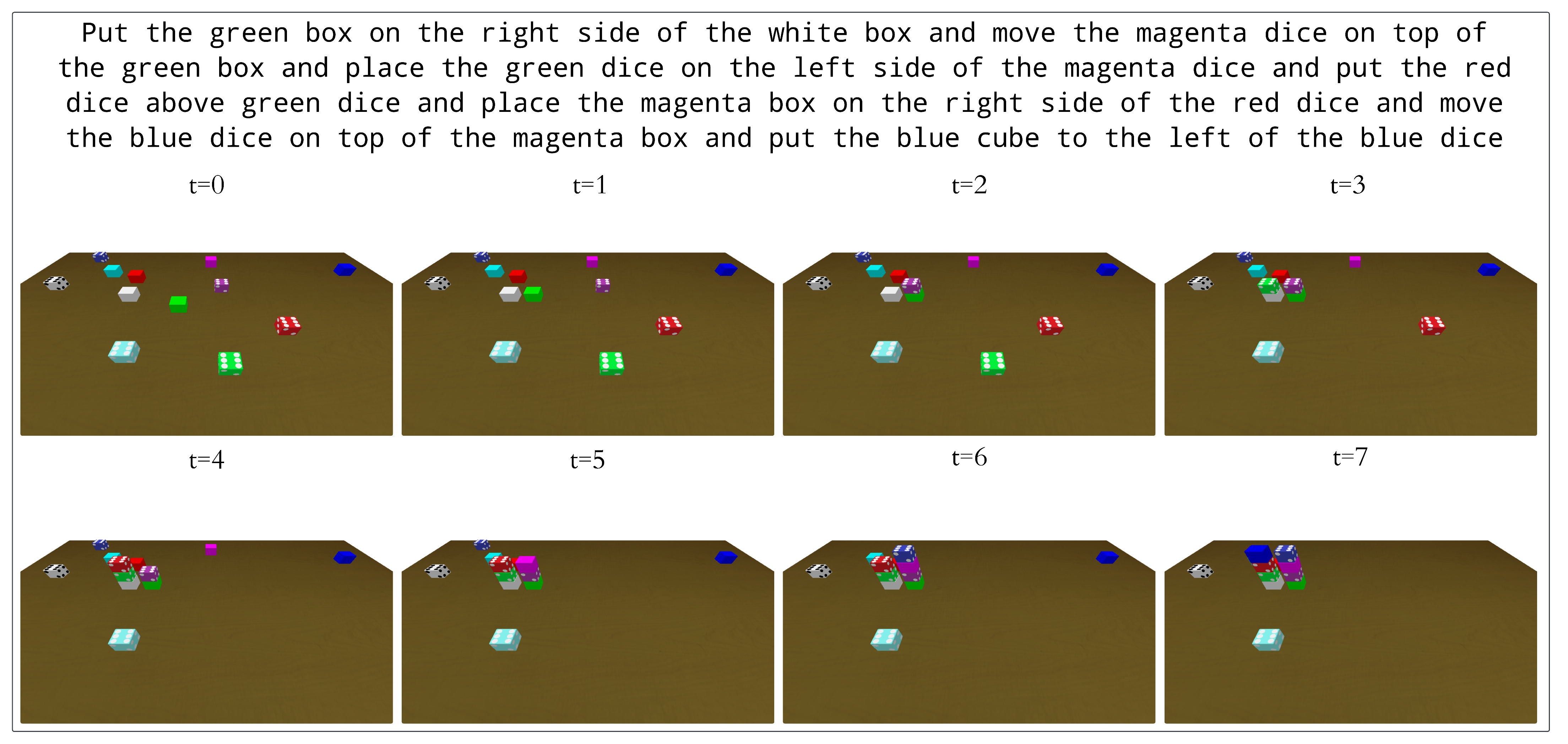}    \caption{\textbf{Example from pre-training dataset.}} 
    \label{fig:pretrain-data}
\end{figure}

\label{app:dataset}
\textbf{Dataset for Pre-training: }
We use 5k examples of constructing \emph{twin-towers} i.e. 2 towers adjacent to each other, for learning semantics of \texttt{move\_head(dir)}, a basic set of visual attributes, reactive policy $\pi_{\text{neural}}$, and neural modules required for grounded planning. The twin towers allow us to learn various action semantics for all possible configurations of blocks in 3D-space (and not being limited to blocks placed directly on table top surface). Since we are not aware of the underlying semantics of \emph{tower} during pre-training phase the corresponding natural language instruction consists of step by step pick and place actions. ~\ref{fig:pretrain-data} gives example demonstrations from this dataset.

\begin{figure}[hbt!]
    \centering
    \includegraphics[width = 1\textwidth]{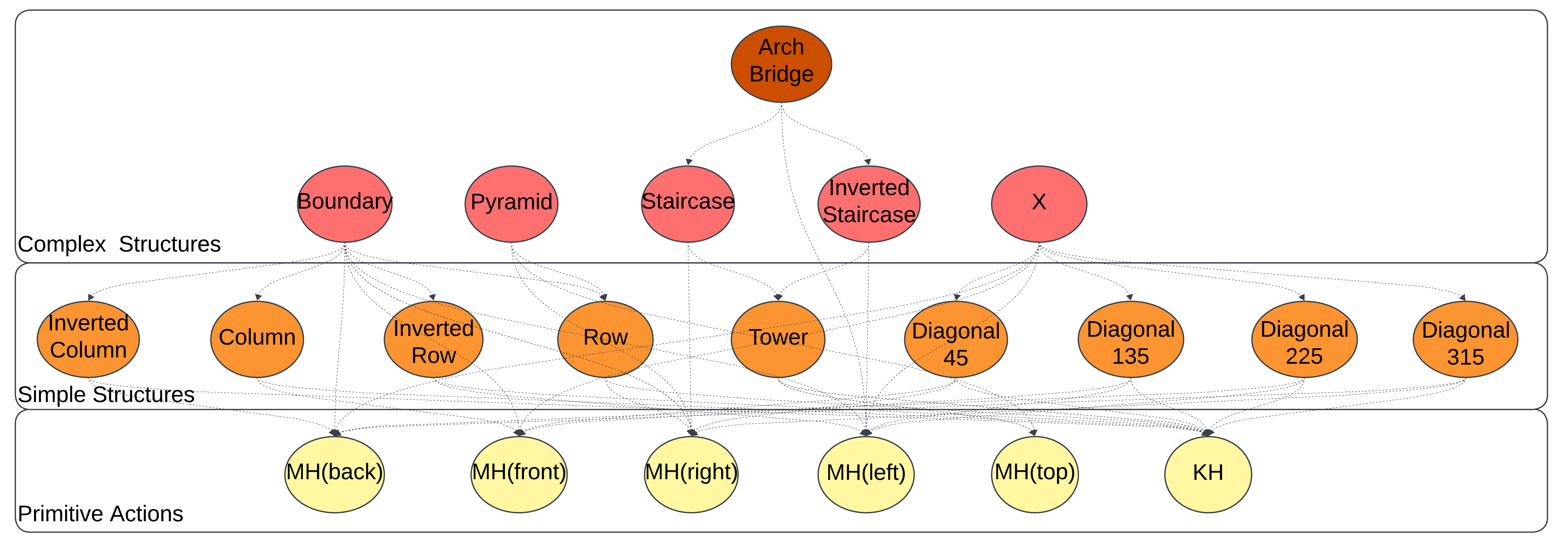}    \caption{\textbf{Hierarchy of the structures/programs.} This diagram shows the hiearchical nature of the structures in our dataset. MH is abbrevation for move\_head and KH is abbreviation for keep\_at\_head} 
    \label{fig:hierarchy}
\end{figure}

\textbf{Dataset for Inductive Structures: }
We learn a variety of structures which we have divided into \emph{Simple} and \emph{Complex} structures. A structure is considered complex if it can be expressed as an inductive composition of simpler structures. As an example, we can express a staircase to be a composition of towers of increasing height. The structures are listed in the Table~\ref{tab:structures}. Figure ~\ref{fig:hierarchy} shows the hierarchical relationship among these structures in the form of a DAG (directed acyclic graph). ~\ref{ic-gt} gives the ground truth program representations for each structure.

\begin{table}[hbt!]
    \begin{minipage}{\textwidth}
        \centering
                \caption{\textbf{Structure Types.} Examples of simple and complex structures considered in this work for the robot to construct. }
        \label{tab:structures}
        \begin{tabular}{|c|c|}
             \hline
             \textbf{Simple Structures} & \textbf{Complex Structures}      \\
             \hline 
             \texttt{Row, Column, Tower} & \texttt{X (cross-shape), Staircase} \\ 
             \texttt{Inverted-Row, Inverted-Column} & \texttt{Inverted-Staircase, Pyramid}  \\
             \texttt{Diagonal-45, Diagonal-135} &  \texttt{Arch-Bridge, Boundary}  \\ 
             \texttt{Diagonal-225, Diagonal-315} & \texttt{}  \\ 
             \hline 
        \end{tabular}
        \vspace{0.10cm}
    \end{minipage}
    \hfill

\end{table}

\section{Prompting Strategy and Examples}
\label{app: prompoting}
%
This section gives various prompting examples for our approach and baselines, along with examples motivating particular design decisions in our approach.

\subsection{{Prompt Example for Task Sketch Generation Stage (\emph{Sketch})}}
\label{app:tsg-prompting}
In order to get a program representation (high level task sketch) of the given natural language instruction, we prompt GPT-4
 with few shot examples in a manner similar to \cite{code-as-policies}. Code segment ~\ref{lst:task_sketch_generation} gives an example of getting the task sketch given the demonstration for constructing a \emph{staircase}. We first import the available primitive operators and functions and also give examples in order to demonstrate the signature of the available primitives(line 1-8). Then we give incontext examples of how to parse various natural language instructions in a program representation(line 10-14). We append to this prompt the instruction for current task(line 16-17).

 \begin{lstlisting}[language=Python, caption={Task Sketch Generation Using GPT-4 (Sketch)}, label={lst:task_sketch_generation}]
# importing the available functions
from visual_operators import filter
from inductive_operators import get_parameters, find_structure

# function signature of the imported functions
filter(color, cube) # filter the objects that are cubes and color ..
get_parameters(structure) # parameters of the structure ...
find_structure(type, description) # finding structure of given type, description

# examples:
# instruction: Find the tower with green cubes
find_structure(type = tower, description = filter(green, cube))
# instruction: Construct a tower of height 3 with yellow cubes
Tower(height = 3, objects = filter(yellow, cubes))

# current task: Construct a staircase of 4 steps using cyan legos

# (GPT-4s output)
Staircase(steps = 4, filter(cyan, legos))
\end{lstlisting}

\subsection{{Prompt Example for Generalizing a sequence of actions/plan to a general program (\emph{Generalize})}}
\label{app:generalize-prompting}
 Code segment ~\ref{lst:generalization} give an example of getting the general Python program from the plan found using MCTS. We first provide a base prompt giving details to GPT-4 about the desired task (line 1-2). Then we give the input arguments and the corresponding output/plan for a given demonstration (line 3-5). We expect the GPT-4 to output the final Python program (line 7-10)\emph{staircase} (line 6-8).

 \begin{lstlisting}[language=Python, caption={Plan to Program using GPT-4 (Generalize)}, label={lst:generalization}]
# Write a general python code which on the given input produces the desired output, do not output anything other than the function description.

# input: n = 4, objects = ObjSet
# output: tower(1, ObjSet), move_head(`right'), tower(2, ObjSet), move_head(`right'), tower(3, ObjSet), move_head(`right'), tower(4, ObjSet)

# Program (GPT-4s output)
def staircase(n, objects):
    for i in range(n):
        tower(i+1, objects)
        move_head('right')
    


\end{lstlisting}

\subsection{Benefit of Estimating Modular/Smaller Plans}
\label{modular-plans}
The below examples demonstrate the benefit of learning new inductive concepts in terms of already acquired inductive concepts (more modular representation). The program ~\ref{lst:pyramid_rows} is obtained through generalization of plan $H_{p}$ that represents \emph{pyramid} in terms of rows of decreasing size.  The generated program generalizes to \emph{pyramid} of different height.
\begin{lstlisting}[language=Python, caption={Pyramid in terms of rows}, label={lst:pyramid_rows}]
# input: n = 3, objects = ObjSet
# output: row(5, ObjSet), move_head('right'), move_head('top'), row(3, ObjSet), move_head('right'), move_head('top'), row(1, ObjSet)

# Program (GPT-4s output)
def pyramid(n, objects)
 for i in range(n, 0, -1):
        # Calculate the number of objects in the current row
        row_count = 2 * i - 1
        
        # Perform actions for the current layer
        row(row_count, objects)
        if i > 1:  # No need to move head after the last row
            move_head('right')
            move_head('top')
\end{lstlisting}
On the other hand program ~\ref{lst:py-prim} which is obtained in an attempt to generalize the plan expressed in terms of primitives is incorrect.

\begin{lstlisting}[language=Python, caption={Pyramid in terms of primitive actions}, label={lst:py-prim}]
# input: n = 3, objects = ObjSet
# output: store_head(), keep_at_head(ObjSet), move_head('right'), keep_at_head(ObjSet), move_head('right'), keep_at_head(ObjSet), move_head('right'), keep_at_head(ObjSet), move_head('right'), keep_at_head(ObjSet), reset_head(), move_head('top'), move_head('right'), store_head(), keep_at_head(ObjSet), move_head('right'), keep_at_head(ObjSet), move_head('right'), keep_at_head(ObjSet), reset_head(), move_head('right'), move_head('top), keep_at_head(ObjSet)

# Program (GPT-4s output)
def pyramid(n, objects)
# Store initial position
    store_head()
    
    # Initial sequence of placing and moving right
    for _ in range(n):
        keep_at_head(objects)
        move_head('right')
    
    # Reset and move to top right position
    reset_head()
    move_head('top')
    move_head('right')
    
    # Store again and repeat placing and moving right
    store_head()
    for _ in range(n):
        keep_at_head(objects)
        move_head('right')
    
    # Reset and adjust final position
    reset_head()
    move_head('right')
    move_head('top')
    keep_at_head(objects)

\end{lstlisting}

\subsection{Comparing Same-length Plans in terms of Generalizability}
Below we illustrate plan to program generalization for 2 different plans which are correct and have same length, but LLMs find it difficult to generalize the second one as compared to first. (Note that row and column of size 1 are equivalent to keep\_at\_head)
 \begin{lstlisting}[language=Python, caption={Plan for tower that can be easily generalized (correct generalization)}, label={lst:correct_gen}]
# input: n = 3, objects = ObjSet
# output: keep_at_head(ObjSet), move_head('top'), keep_at_head(ObjSet), move_head('top'),  keep_at_head(ObjSet)


# Program
def tower(n, objects):
    for _ in range(n):
            keep_at_head(objects)
            move_head('top')
\end{lstlisting}

 \begin{lstlisting}[language=Python, caption={Plan for tower that is difficult to generalize (Incorrect generalization)}\label{lst:incorrect_gen}]
# input: n = 3, objects = ObjSet
# output: row(1, ObjSet), move_head('top'), keep_at_head(1, ObjSet), move_head('top'),  column(1, ObjSet)


# Program 
def tower(n, objects):
    for i in range(1, n + 1):
            row(i, objects)
            move_head('top')
            keep_at_head(objects)
            move_head('top')
            if i < n:
                column(i, objects)
                move_head('top')
\end{lstlisting}

\subsection{Generalizing via Multiple Demonstrations}
\label{gen-mul-plan}
Given multiple demonstrations we independently find task sketch and corresponding grounded plans for each demonstration. These are further given to GPT-4 for generalization. Code segment ~\ref{lst:mul_plan_gen} gives an example of getting a single Python program from multiple demonstrations. Note that we explicitly prompt the LLM that some of the grounded plans might be incorrect (which may lead to more robust generalization in case of noisy demonstrations). 

\begin{lstlisting}[language=Python, caption={Generalizing through multiple plans}, label={lst:mul_plan_gen}]

# Function Call: wor(height = 3, objects = ObjSet_1)
# Execution: keep_at_head(obj = ObjSet_1), move_head(dir = right), keep_at_head(obj = ObjSet_1), move_head(dir = right), keep_at_head(obj = ObjSet_1), 
# Function Call: wor(height = 3, objects = ObjSet_1)
# Execution: keep_at_head(obj = ObjSet_1), move_head(dir = right), keep_at_head(obj = ObjSet_1), move_head(dir = right), keep_at_head(obj = ObjSet_1), 
# Function Call: wor(height = 3, objects = ObjSet_1)
# Execution: column(size=1, obj = ObjSet_1), move_head(dir = right), keep_at_head(obj = ObjSet_1), move_head(dir = right), keep_at_head(obj = ObjSet_1), 

#Write the function definition, which generalizes the above executions. Note that some of the executions can be partially wrong. 
```python
def wor(height, objects):
```

GPT-4s Output .....
\end{lstlisting}

\subsection{Prompt Examples for Learning Programs using LLM/VLM Models}
Below we describe the prompting methodologies for learning programs through LLM/VLM models. Note that although the prompt examples described below are for the case of learning novel structure from 1 demonstration, we use 3 demonstration per novel structure in our main results (for both our approach and LLM/VLM baseline).
\label{vlm-llm-baseline}

\textbf{LLM/GPT-4} Code segment \ref{lst:llm_baseline_code} depicts our prompting methodology given a demonstration for a new concept \emph{tower}. For this baseline we aim to check demonstration following and spatial reasoning abilities of LLMs (GPT-4). We provide supervision of the intermediate scenes by using tokenized spatial relations between objects in the scene (Given in the form of Scene = [right(1, 0) ...]). We further assume that only those objects that are required to perform the task are present in the scene (no distractor objects). For every structure (that needs to be learned at time t) we give LLM a prompt providing in-context example on how to generalize (line 19-35), the set of primitive operators (line 4) available and the set of structures learnt/present in library (till time t-1) (line 5-18). Finally we append to this prompt the expected declaration (arguments and keywords arguments) of the inductive concept that is to be learnt along with the spatial relations for each scene of the given demonstration (36-51). Note that we assume absence of distractor objects for this baseline.

\begin{lstlisting}[language=Python, caption={Prompting Strategy for LLM baselines (GPT-4)}, label={lst:llm_baseline_code}]
# Consider a block world domain ...... Given a structure creation task along with intermediate scnes complete a general Python function for it. The function should be in terms of primitive operators and already learnt structures that are present in the program library. Enclose the function within backtick (```)

primitive_operators = [keep_at_head, move_head ..]
# this would be our program library
learnt_structures = {
    "row": {
            "program_tree": 
            '''
            def row(size, objects): 
                for i in range(size):
                    keep_at_head(obj = objects)
                    move_head(dir = 'right')
            ''',
        },

    ......
}
# the example task
Example task:- Place all the objects to the right of each other.
Final state :- [right(1, 0), right(2, 1), right(3, 2), right(4, 3)]
Intermediate scenes :- 
Scene 0 = []
Scene 1 = []
Scene 2 = [right(1, 0)]
Scene 3 = [right(1, 0), right(2, 1)]
Scene 4 = [right(1, 0), right(2, 1), right(3, 2)]
Scene 5 = [right(1, 0), right(2, 1), right(3, 2), right(4, 3)]
Python function :-
```python
def placing_all_right(objects):
    for i in range(len(objects)):
        keep_at_head(objects) # select one object from the objects set and keep the head at this location
        move_head(dir = 'right') # move the head to the right of the previous position
```
# The current task for which program needs to be found
Current task:- Construct a tower of size 6.
Final state :- [top(1, 0), top(2, 1), top(3, 2), top(4, 3), top(5, 4)]
Intermediate scenes :- 
Scene 0 = []
Scene 1 = []
Scene 2 = [top(1, 0)]
Scene 3 = [top(1, 0), top(2, 1)]
Scene 4 = [top(1, 0), top(2, 1), top(3, 2)]
Scene 5 = [top(1, 0), top(2, 1), top(3, 2), top(4, 3)]
Scene 6 = [top(1, 0), top(2, 1), top(3, 2), top(4, 3), top(5, 4)]
Python function :-
```python
def tower(size, objects):
??
```
\end{lstlisting}

\begin{figure}[hbt!]
    \centering
    \includegraphics[width=0.99\textwidth]{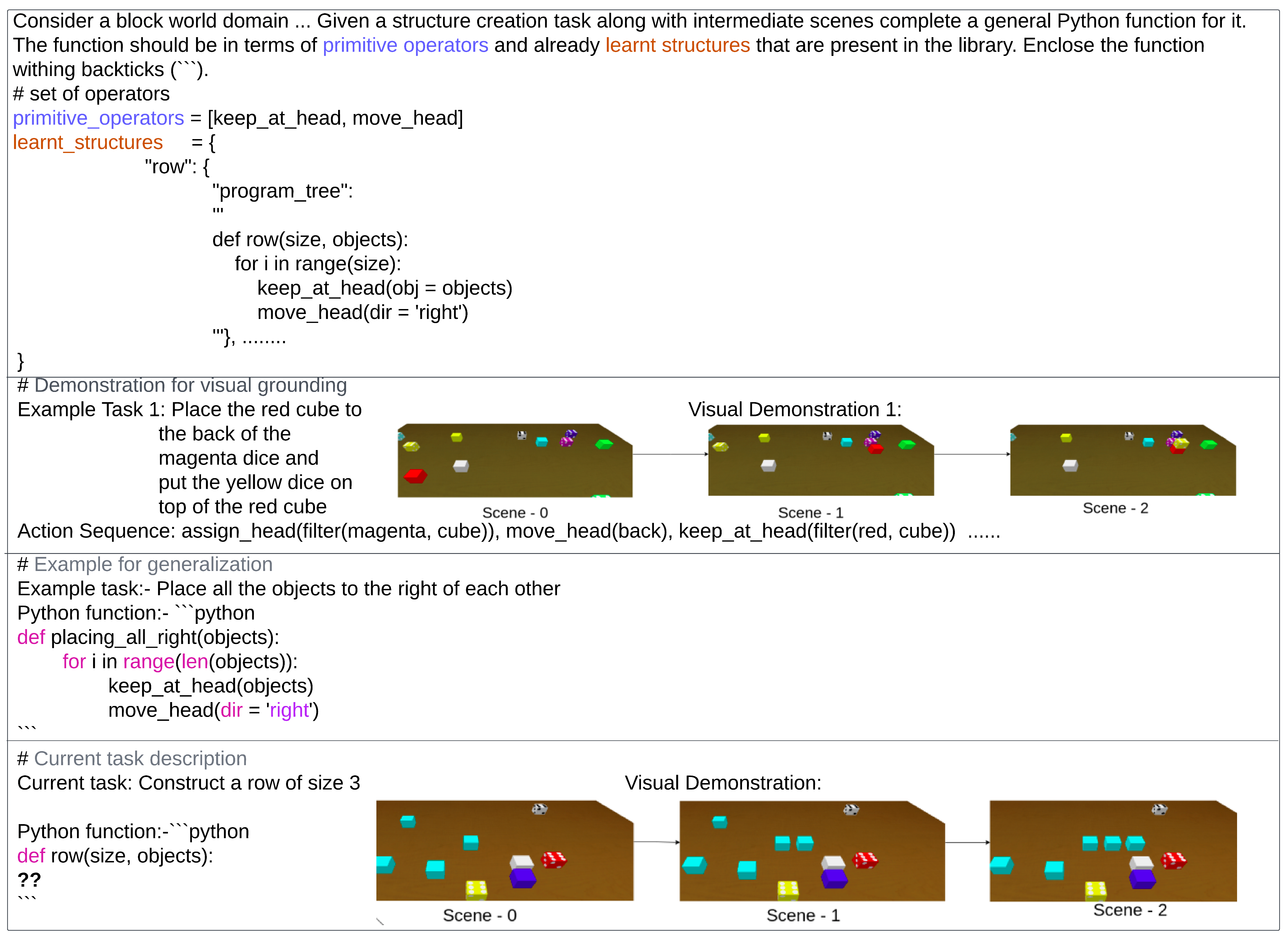}
    \caption{
    \textbf{Prompt example for VLM-baseline.} Figure shows the prompting strategy for GPT-4-V that includes the primitive actions, the example tasks for visual grounding, example of writing generalizable Python functions and the demonstration frames.}
    \label{fig:vlm_baseline}
\end{figure}

\label{vlm-baseline}
\textbf{VLM/GPT-4-V} Unlike LLM, VLMs have the abilities to process the demonstration as a sequence of visual frames. Therefore rather than providing the symbolic spatial relations between every scene we instead directly provide all the intermediate scenes for the given demonstration. Further we also relax the assumption that there are no distractor objects. As shown in figure \ref{fig:vlm_baseline} We first give information about the set of primitive operators and the structures that we have already learnt (library of concepts). In order to visually ground the semantics of our primitive actions we give 3 example tasks (natural language instruction and intermediate scenes) that do not directly refer to any structure, along with corresponding sequence of actions taken (\emph{\# Demonstration for visual grounding}). We further provide another example (without scenes) demonstrating how to write generalizable Python function for a given task using our operators (\emph{\# Example for generalization}). Finally we give the natural language instruction and corresponding scenes for the current task along with signature of the program to be learnt (\emph{\# Current task description}).

\newpage
\section{Supplementary Results}
\textbf{Out-of-Distribution Performance:}
\begin{table}[hbt!]
\centering
\caption{Out-of-Distribution Performance (mean $\pm$ std-error)}
\label{table:out-dist-std}
\begin{tabular}{@{}lcccc@{}}
\toprule
\multirow{2}{*}{Model} & \multicolumn{2}{c}{Simple} & \multicolumn{2}{c}{Complex} \\
\cmidrule(lr){2-3} \cmidrule(lr){4-5}
& IoU & MSE & IoU & MSE \\
\midrule
SPG(Ours) & \textbf{0.892} $\pm$ 0.065 & \textbf{4.386e-4} $\pm$ 2.387e-4 & \textbf{0.804} $\pm$ 0.025 & \textbf{0.001} $\pm$ 5.391e-4 \\
GPT-4 & 0.776 $\pm$ 0.023 & 0.006 $\pm$ 0.001 & 0.131 $\pm$ 0.019 & 0.019 $\pm$ 1.498e-3 \\
GPT-4V & 0.575 $\pm$ 0.026 & 0.013 $\pm$ 0.001 & 0.290 $\pm$ 0.016 & 0.011 $\pm$ 1.316e-3 \\
SD & 0.236 $\pm$ 0.005 & 0.006 $\pm$ 7.495e-4 & 0.150 $\pm$ 0.011 & 0.011 $\pm$ 2.860e-3 \\
SD+G & 0.273 $\pm$ 0.004 & 0.006 $\pm$ 6.180e-4 & 0.154 $\pm$ 0.010 & 0.014 $\pm$ 2.958e-3 \\
\bottomrule
\end{tabular}
\end{table}


\textbf{Performance \textit{Dataset II} (i.e. name reversed evaluation):} Table ~\ref{table:in-dis-nr} gives the corresponding program accuracies, while Table ~\ref{table:in-dis-nr-std} give the corresponding IoU/MSE metrics along with standard errors.


\begin{table}
\captionsetup{font=small}
\caption{Performance on \emph{Dataset II}(Names Reversed)}
\label{table:in-dis-nr-std}
\centering
\resizebox{0.8\textwidth}{!}{ 
\begin{tabular}{l c c c c}
\toprule
\multirow{2}{*}{\textbf{Model}} & \multicolumn{2}{c}{\textbf{Simple}} & \multicolumn{2}{c}{\textbf{Complex}} \\
\cmidrule(lr){2-3} \cmidrule(lr){4-5}
& \textbf{IoU} & \textbf{MSE (1e-3)} & \textbf{IoU} & \textbf{MSE (1e-3)} \\
\midrule
SPG(Ours) & \textbf{0.86} $\pm$ 0.03 & \textbf{1.74} $\pm$ 0.45 & \textbf{0.78} $\pm$ 0.02 & \textbf{3.93} $\pm$ 1.09 \\
GPT-4 & 0.78 $\pm$ 0.03 & 3.16 $\pm$ 0.51 & 0.00 $\pm$ 0.00 & 22.73 $\pm$ 1.48 \\
GPT-4V & 0.71 $\pm$ 0.01 & 3.92 $\pm$ 0.45 & 0.09 $\pm$ 0.02 & 21.29 $\pm$ 1.59 \\
\bottomrule
\end{tabular}
}
\vspace{-3mm}
\end{table}

\subsection{Qualitative Comparison between Purely-neural (Struct-Diff+Grounder) vs. Ours(SPG)}
\begin{figure}[hbt!]
    \centering
    \includegraphics[width =0.6\textwidth]{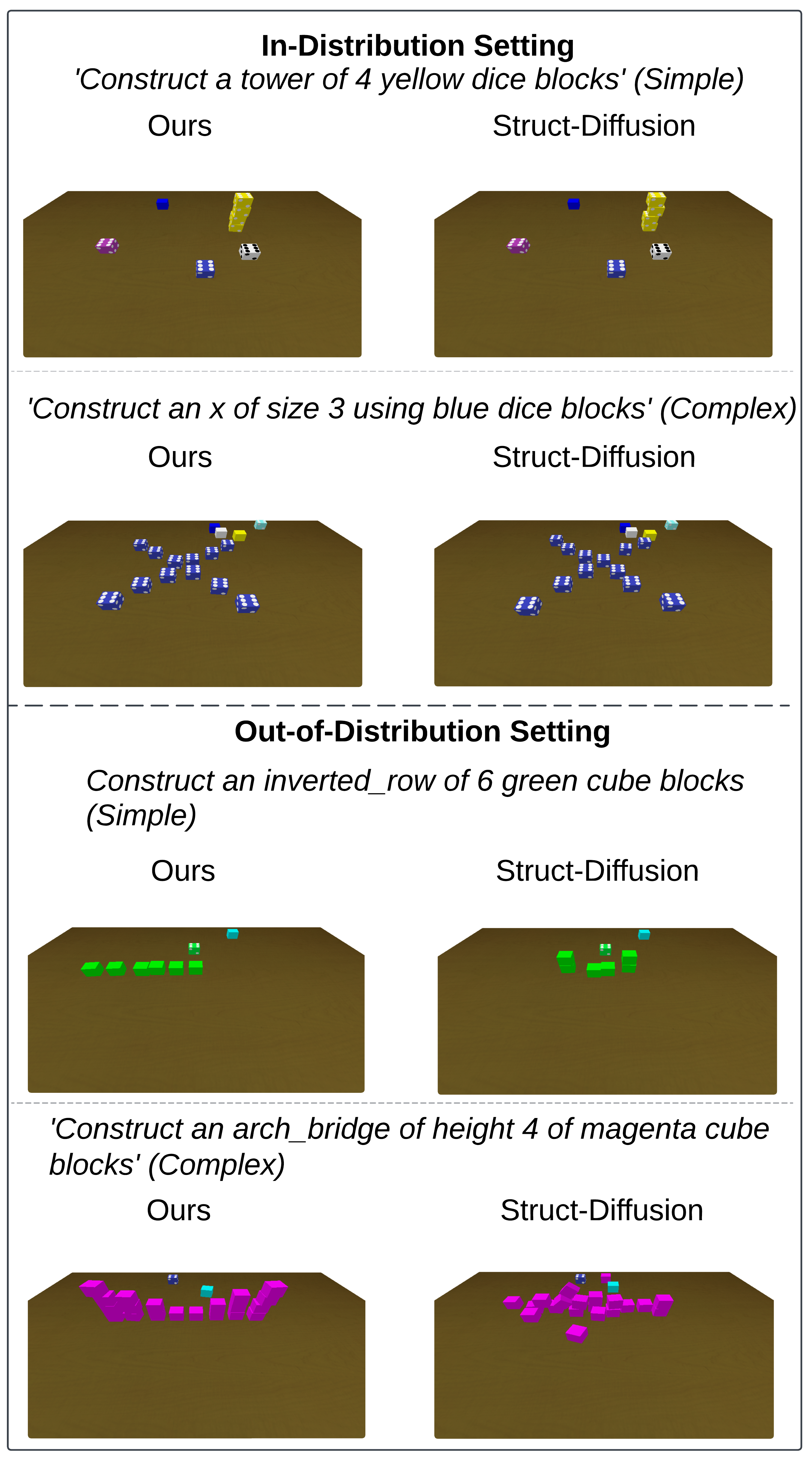}
    \caption{\textbf{Structure creation comparison between SPG(Ours), and Struct-Diff+Grounder}} 
    \label{fig:struct-creation-qual}
\end{figure}

Figure ~\ref{fig:struct-creation-qual} gives compares the qualitative results for our approach against Struct-Diffusion with grounder on both in-distribution, \emph{Dataset I} and out-of-distribution (larger size), \emph{Dataset III}. In in-distribution setting the our method performs slightly better in terms of structure creation for both simple and complex structures, but the difference is not significant. However for out-of-distribution setting structures created by our approach are much better than those created by Struct-Diffusion+Grounder. Further for this setting structure creation by Struct-Diffusion seems to be much worse for complex structures than simple ones.

\subsection{Continual Learning of Neural Concepts}
\label{app:continual-neural}

\begin{figure}[H] 
    \vspace{-3mm}
    \centering
    \caption{Disentanglement. \footnotesize{The acquisition of new visual concepts. Plot shows an increase in the likelihood of correct grounding of an object referenced with a new neural concept (\emph{chocolate} color) with training iterations.}}
    \resizebox{0.40\textwidth}{!}{ 
        \includegraphics[width=0.80\linewidth, height=0.40\linewidth]{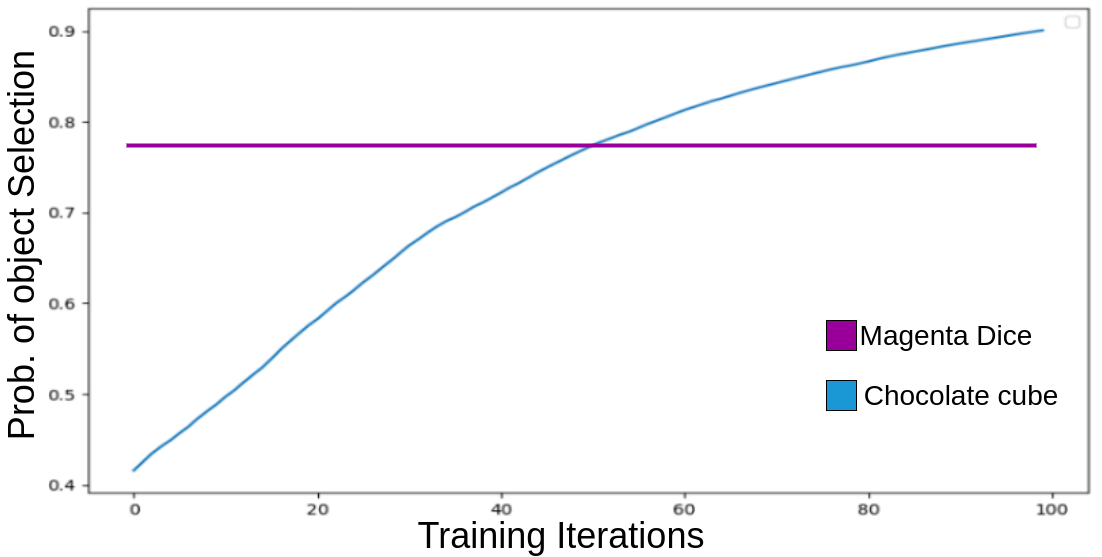}
    }
\label{fig:gumbel}
    \vspace{-3mm}
\end{figure}

Having a disentangled representations allows us to (i) intersperse learning of new visual attributes with learning of new inductive concepts (ii) avoid catastrophic forgetting of already learnt attributes. For example, the model can learn the \emph{chocolate} from an instruction \quotes{construct a tower using chocolate blocks of size 4}, even if it has not seen the color in the pretraining phase. Because of our modular architecture, we can learn the color as a new embedding in the space of visual attributes. The plot in Fig. ~\ref{fig:gumbel} demonstrates the benefit of having such disentangled representations. As the training proceeds the probability of being able to select the \emph{chocolate} blocks when required increases with time, while keeping the ability of selecting a magenta colored object (when required) remains the same. 

\begin{figure}[hbt!]
    \centering
    \includegraphics[width =\textwidth]{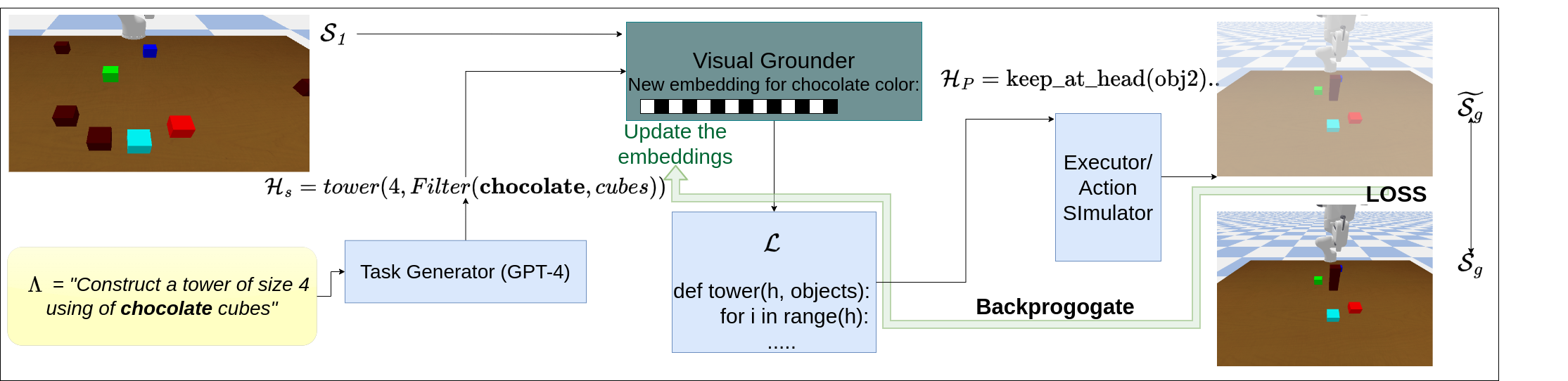}
    \caption{\textbf{Continual learning of visual primitives}} 
    \label{fig:gumbel-method}
\end{figure}

Given demonstration for the task \quotes{Construct a tower of height 4 using \emph{chocolate} cubes}, we would like to learn the neural embedding for the unknown color \emph{chocolate} (where we assume that tower has already been learnt and stored in the library $\mathcal{L}$). First the instruction is converted into corresponding plan sketch $\mathcal{H}_{S}$ = tower(4, Filter(\textbf{chocolate}, cubes)), which is passed to the visual grounder. The grounder detects the presence of an unknown attribute \emph{chocolate} as an argument to filter, and randomly initializes a new neural embedding for it. Using this new embedding along with the already present embedding of cube and the visual features found through ResNet-34, the quasi-symbolic executor outputs a grounded task-sketch. The executor executes the grounded task-sketch by getting the semantics of the underlying function i.e. tower from the library $\mathcal{L}$. MSE+IoU loss computed over the final scene obtained and the expected final scene is backpropogated through the network. Note that during backpropogation all the neural modules (action semantics, visual attributes, ResNet-34) are frozen, except for the newly initialized embedding for \emph{chocolate}. For the purpose of differentiable sampling during tower construction we use gumbel-softmax \cite{gumbel} with masking. Figure ~\ref{fig:gumbel-method} illustrates our approach.

\subsection{Details for Inference on Novel Tasks using an LLM}
\label{details-hri}

Below we show the \cite{code-as-policies} inspired prompting methodology that we use to get the executable code corresponding to a language specified manipulation task.
We initially begin by importing the helper functions, spatial direction, primitive functions, and learnt inductive concepts/structures (line 5-11). Then we give few examples for how to use and compose the various functions for different tasks (line 16-83). Finally we give the instruction of current task, and expect GPT-4 to output the corresponding executable code. 

\begin{lstlisting}[language=Python, caption={Prompting method for the task of constructing tower of white cubes to the same height as existing tower of green die}, label={def:task-same-size}]
# Given a task you have to provide Python code for executing the task

# importing available functions

from spatial_directions import top, front, back, left, right

from primitives import assign_head, move_head, keep_at_head 
# HEAD is a imaginary pointer keeping track of the current spatial location in consideration

from helpers import find_size, filter
from structures import row, column, tower


#function signature of the imported functions
# finds all the objects with the given color and shape, returns a mask denoting the probability of object selection
filter(color, shape)

# finds the size of the structure struct_name that is formed with objects of the given type, returns the size of the structure (whose type is integer), arguments for this should be provided as kwargs
find_size(struct_name = str_name, objects = ObjSet)

# assigns the head to the location of the object
assign_head(at_obj_loc)

# moves the head in the given dir
move_head(dir)

# keeps the object obj at the head
keep_at_head(obj)


#Examples:
#Instruction: Move the green block to the left of the red dice
assign_head(at_obj_loc = filter(red, dice))
move_head(left)
keep_at_head(obj = filter(green, cube))

# Instruction: Find the size of the tower made of yellow legos
find_size(struct_name = tower, objects = filter(yellow, lego))

#Instruction: Find the size of the row made of orange cubes
find_size(struct_name = row, objects = filter(orange, cube))

# Instruction: Find the size of the column made of cyan cubes
find_size(struct_name = column, objects = filter(cyan, cube))

# Instruction: Move the green block to the left of the red dice and the yellow block to the top of the green block
assign_head(at_obj_loc = filter(red, dice))
move_head(left)
keep_at_head(obj = filter(green, cube))
assign_head(at_obj_loc = filter(green, cube))
move_head(top)
keep_at_head(obj = filter(yellow, cube))

# Instruction: Construct a row of green legos of length 3 to the right of the blue block
assign_head(at_obj_loc = filter(blue, block))
move_head(right)
row(length = 3, objects = filter(green, legos))


# Instruction: Construct a tower of size 3 using red cubes
tower(height = 3, objects = filter(red, cube))

# Instruction: Construct a row of size 5 using blue legos
row(length = 5, objects = filter(blue, lego))

# Instruction: Construct a column of size 6 using green die
column(length = 6, objects = filter(green, dice))

# Instruction: Place 3 green blocks so that one block is to the right of the other
green_blocks = filter(green, block)
keep_at_head(green_blocks)
move_head(right)
keep_at_head(green_blocks)
move_head(right)
keep_at_head(green_blocks)

# Instruction: Place 3 red legos on top of one another
red_legos = filter(red, lego)
keep_at_head(red_legos)
move_head(top)
keep_at_head(red_legos)
move_head(top)
keep_at_head(red_legos)


# CURRENT TASK 
# Instruction: Construct tower of white cubes to the same height as
# existing tower of green die

# GPT-4s output
#First, we have to find the height of the tower of green dice, 
#then construct a tower of white cubes of the same size

tower_size = find_size(struct_name = tower, objects = filter(green, die))
tower(height = tower_size , objects = filter(white, cube))
        
\end{lstlisting}

\begin{lstlisting}[language=Python, caption={Prompting method for the task of constructing tower of alternating red and blue cubes}, label={def:task-alternate}]
# Given a task you have to provide Python code for executing the task

# importing available functions
from spatial_directions import top, front, back, left, right
from primitives import assign_head, move_head, keep_at_head 
# HEAD is a imaginary pointer keeping track of the current spatial location in consideration
from helpers import find_size, filter
from structures import row, column, tower
#function signature of the imported functions
# finds all the objects with the given color and shape, returns a mask denoting the probability of object selection
filter(color, shape)
# finds the size of the structure struct_name that is formed with objects of the given type, returns the size of the structure (whose type is integer), arguments for this should be provided as kwargs
find_size(struct_name = str_name, objects = ObjSet)
# assigns the head to the location of the object
assign_head(at_obj_loc)
# moves the head in the given dir
move_head(dir)
# keeps the object obj at the head
keep_at_head(obj)

#Examples:
#Instruction: Move the green block to the left of the red dice
assign_head(at_obj_loc = filter(red, dice))
move_head(left)
keep_at_head(obj = filter(green, cube))

# Instruction: Find the size of the tower made of yellow legos
find_size(struct_name = tower, objects = filter(yellow, lego))

#Instruction: Find the size of the row made of orange cubes
find_size(struct_name = row, objects = filter(orange, cube))

# Instruction: Find the size of the column made of cyan cubes
find_size(struct_name = column, objects = filter(cyan, cube))

# Instruction: Move the green block to the left of the red dice and the yellow block to the top of the green block
assign_head(at_obj_loc = filter(red, dice))
move_head(left)
keep_at_head(obj = filter(green, cube))
assign_head(at_obj_loc = filter(green, cube))
move_head(top)
keep_at_head(obj = filter(yellow, cube))

# Instruction: Construct a row of green legos of length 3 to the right of the blue block
assign_head(at_obj_loc = filter(blue, block))
move_head(right)
row(length = 3, objects = filter(green, legos))


# Instruction: Construct a tower of size 3 using red cubes
tower(height = 3, objects = filter(red, cube))

# Instruction: Construct a row of size 5 using blue legos
row(length = 5, objects = filter(blue, lego))

# Instruction: Construct a column of size 6 using green die
column(length = 6, objects = filter(green, dice))

# Instruction: Place 3 green blocks so that one block is to the right of the other
green_blocks = filter(green, block)
keep_at_head(green_blocks)
move_head(right)
keep_at_head(green_blocks)
move_head(right)
keep_at_head(green_blocks)

# Instruction: Place 3 red legos on top of one another
red_legos = filter(red, lego)
keep_at_head(red_legos)
move_head(top)
keep_at_head(red_legos)
move_head(top)
keep_at_head(red_legos)


# CURRENT TASK 
# Instruction: Construct a tower of height 6 using red and blue blocks that are alternating

# GPT-4s output
# Python code:

# Define the red and blue blocks
red_blocks = filter('red', 'block')
blue_blocks = filter('blue', 'block')

# Start at the bottom and alternate building the tower
for i in range(6):
    if i%2 == 0: # if the stack position is even
        keep_at_head(red_blocks)
    else: # if the stack position is odd
        keep_at_head(blue_blocks)
    if i != 5:  # if not at the top of the tower
        move_head(top)
        
\end{lstlisting}

To find the size of a given structure in the given scene we define the function \texttt{find\_size}, which takes the name of structure, all the objects in the initial scene, mask of the objects (a distribution over the objects based on the attributes), and the initial state. (we assume that this function has access to the semantics of all the concepts learnt so far, through a transition function). Algorithm ~\ref{alg:find-size} gives the pseudocode for the function find\_structure. Below we provide a brief explanation for it.
\begin{enumerate}
    \item First we assign our head to every block in the available blocks (line 6)
    \item Then we begin constructing/visualizing the corresponding structure from that block beginning with a size of 1. (line 7)
    \item For each structure created/visualized we compare the blocks moved for the structure creation with corresponding blocks originally present in the scene, and perform a matching between these blocks and a subset of the blocks originally present (line 11-26).
    \item If we are able to find a mapping for each moved block, such that each mapped pair has an IoU greater than a threshold, we increase the next potential size to test by 1 (line 26-27).
    \item The final size is the size corresponding to 2nd last iteration, before termination (line 28).
    \item We return the maximum of all the possible structures that are found (line 34)
\end{enumerate}

\subsection{Details on MCTS Variants for Plan Search}
\label{app:diff-plan-search-details}
Here we provide the details for 3 different plan search methods, that search over the space $\mathcal{A} = \mathcal{A}_{c} \cup \mathcal{A}_{p}$
\begin{itemize}
    \item MCTS+\emph{L}+\emph{P}: This is the approach that we describe in section 4.2. For every concept say \texttt{Tower} $\in \mathcal{L}$ we have a corresponding set of macro action say \texttt{Make\_Tower(3, objects)} (\emph{L}). Further we use a neural pruner $\pi_{\text{neural}}$ that outputs a primitive action $a_{p}^*$ (given current state and next expected state). We only consider the actions $\mathcal{A}_{c} \cup \{a_{p}^*\}$ during our search from the given state. This helps to reduce the effective branching factor and allows to search longer length plans within the same computational budget.
    \item MCTS+\emph{L}-\emph{P}: Here we do not use the reactive policy, therefore the branching factor for every node becomes $\mathcal{A}$ = $\mathcal{A}_{c} \cup \mathcal{A}_{p}$.
    \item MCTS-\emph{L}+\emph{P}: We only search among the space of primitive actions i.e $\mathcal{A}_{p}$. Given a state $\tilde{s}_{t}$ and corresponding next expected state $s_{t+1}$ we greedily pick the action $a_{p}^* = \pi_{\text{neural}}(\tilde{s}_{t}, s_{t+1})$. This method is much more faster than the previous 2 methods as there is no explicit search. However the corresponding policy is trained only to output an action $a \in \mathcal{A}_{p}$ and lacks the ability to output modular plans composed of macro actions such as \texttt{Make\_Tower(3, objects)} $\in \mathcal{A}_{c}$ (the action space $\mathcal{A}_{c}$ is increasing with time and the architecture of network needs to be changed accordingly). As a result the plans found are not modular and difficult to generalize. Further, the reactive policy is not trained to output \texttt{reset\_head(), store\_head()} as additional annotated data is required in order to train a classifier over them. This further decreases the space of grounded plans (and therefore corresponding generic programs) such a policy can represent. Training a reactive policy that can handle actions such as \texttt{reset\_head()} and an action space $\mathcal{A}_{c}$ that grows with time is part of future work.
\end{itemize}

\subsection{Goal-conditioned Planning with Learnt Concepts}
\label{app:planning-details}

\begin{wrapfigure}{hr}{0.4\textwidth} 
  \centering
  \includegraphics[width=0.4\textwidth]{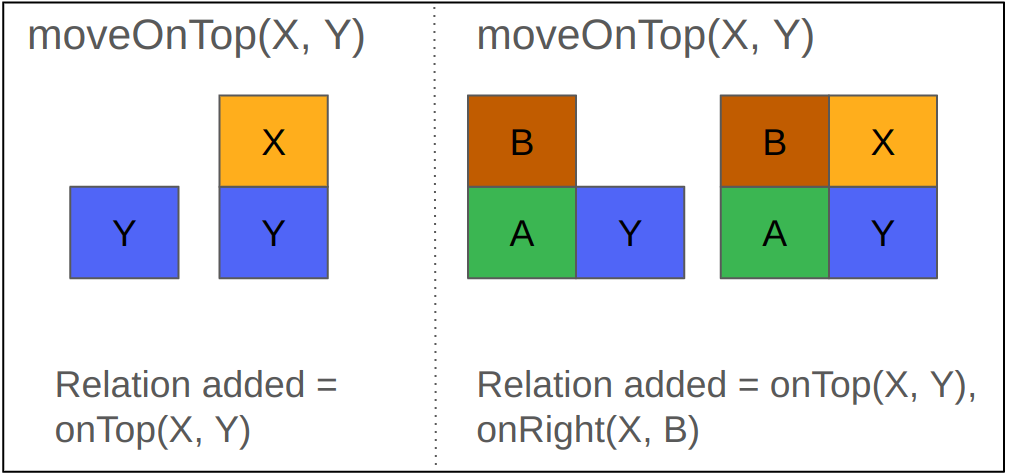} 
  \caption{
    \textbf{Difficult to encode post-conditions}. Illustration of a domain where encoding a PDDL for direct planning is challenging. }
    \label{fig:pddl_hard}
\end{wrapfigure}

\textbf{Why is it difficult to hand encode a PDDL for our domain?}
\label{app:pddl-hard-subsec}
Most of the PDDL description of blocks world assume actions involving only the spatial relation \texttt{onTop}, which limits their applicability to describing different structures like \emph{row} that need spatial relations like onRight. Further a single action might lead to varied effects/post-conditions based on the initial state.~\ref{fig:pddl_hard} gives 2 example of the same action \texttt{moveOnTop(A, B)} which would end up giving adding different number of spatial relations.

\textbf{Approach Overview.}
Given an instruction $\Lambda$ = \quotes{Construct a staircase of magenta die having 3 steps}, we first convert it into corresponding grounded task sketch $H_S^*$ = \texttt{staircase(3, [3, 2, 1, $\cdots$])}. Executing the corresponding program of staircase (by getting the semantics from the library $\mathcal{L}$) on the desired objects we get the expected final scene $S_{f}'$ in bounding box space. The initial scene $S_{i}$ and expected final scene $S_{f}'$ are converted into scene graph $SG_{i}'$ and $SG_{f}'$ (described in ~\ref{app:scene-graph-gen-subsec}). The relations between the task relevant objects in $SG_{f}'$ act as propositions/relations for goal check and the initial scene graph act as the initial state. Then a neuro-symbolic planner is used to obtain the optimal plan from the start state to a state that satisfies the goal. Below we also detail different aspects of the approach.

\textbf{Scene-graph Extraction.}
\label{app:scene-graph-gen-subsec}
~\ref{scene-graph-algo} gives the algorithm used for generating scene graph from a given scene (set of bounding boxes). Suppose we need to check whether there exists a relation of the form $(i, j, direction)$ i.e. block $i$ is in the direction $direction$ of block $j$, in a given scene. We first initialize the head at the position/bounding-box of block $j$ (line 7). Then we move the head in direction $direction$ (line 9). We claim that the relationship would exists if bounding box for block $i$ has IoU $>$ 0.75 with the predicted\_head (line 10-12).

    

\textbf{Pre-conditions.}
We define the following 2 preconditions (and learn their grounding) in order to ensure that the generated plans are physically possible.

\begin{figure}[hbt!]
\centering
\includegraphics[width=0.9\textwidth]{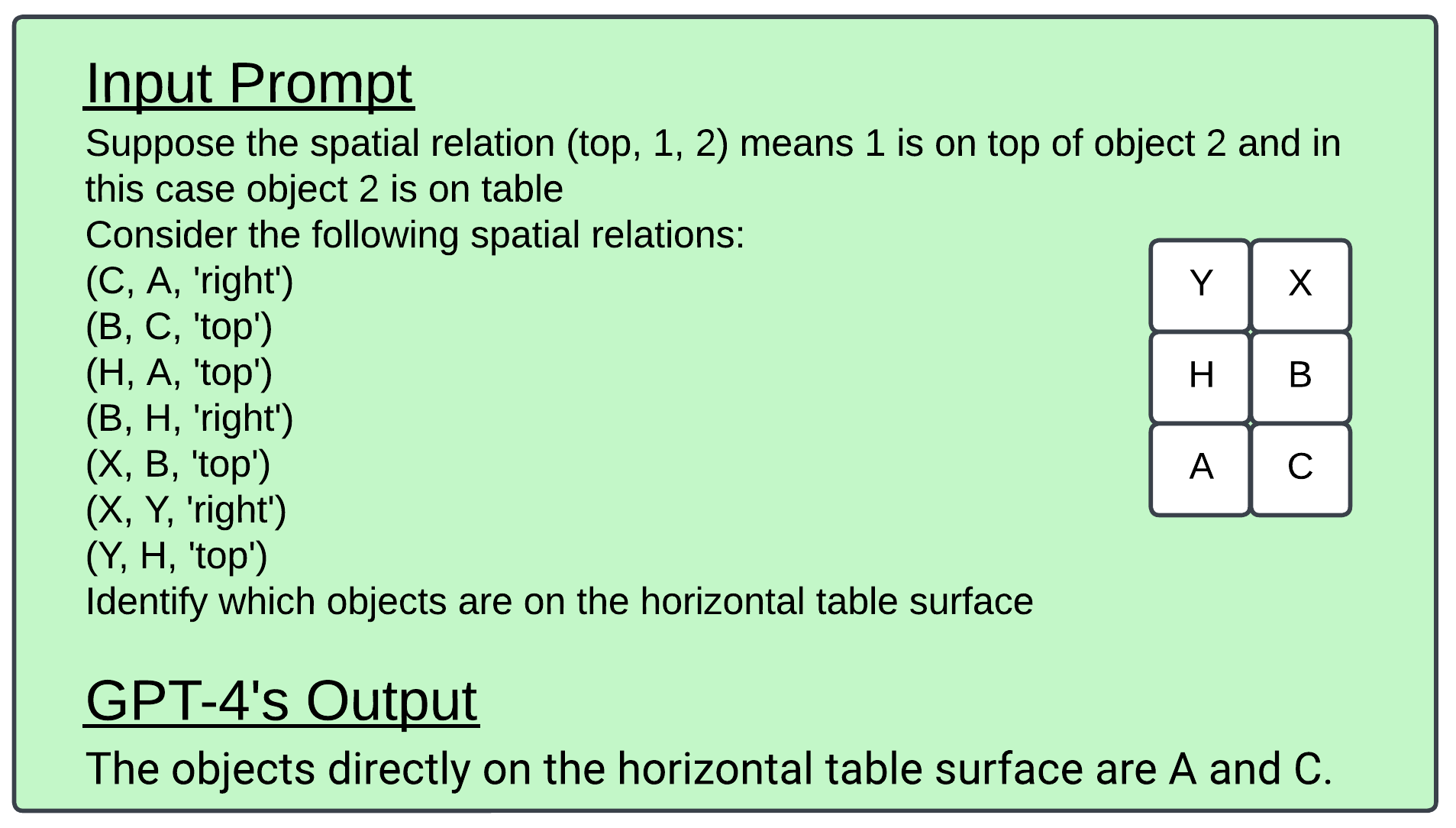}
\caption{
\textbf{Method to get annotation for training on-table classifier.}}
\label{fig:on-table-det}
\end{figure}

\begin{enumerate}[noitemsep]
    \item \texttt{is-clear(blk, dir):} We need to check whether a block blk has some free space in direction dir. For this we simply move the head in the direction dir with respect to the block blk. If the resulting position of head has 0 overlap with bounding boxes of all the other objects the predicate is True otherwise False.

    \item \texttt{will-not-be-floating(pred\_loc):} We need to check whether the resultant/predicted location of an object on taking an action would be dynamically stable or not. The location would be stable if either it is on top of some already placed object or it is on the table surface. The former can be checked through the resultant scene graph itself (that is obtained by applying the algorithm ~\ref{scene-graph-algo}), while for the later we train an \emph{on-table} classifier. This would take as input a bounding box and predict whether the box is on the table or not. For training this classifier we use the dataset of pretraining phase. The blockwise positive and negative sample annotation can be done automatically by giving GPT-4 the corresponding scene graph and then querying which objects are on the table and which aren't. ~\ref{fig:on-table-det} gives an example. (Though we have not taken this approach for the complete dataset of 5k samples due to high cost).
\end{enumerate}

\textbf{Actions.}
We define the following two types actions
\begin{enumerate}[noitemsep]
    \item \texttt{place-random(blk):} To place a block at a random free position on the table. For this we train a generative model (VAE) which learns the underlying distribution of bounding boxes for all the blocks that lie on the table. Given a scene we would sample position (bounding box) from this until we get a position that is not overlapping with the existing blocks in the scene. For training the VAE we assume that for every demonstration in the pretraining data, the first scene has all the objects randomly placed on the table (we could have also used the positive examples used for training on-table classifier). 
    
    \item \texttt{move(rel, blk1, blk2):} This action corresponds to moving the blk1 in the direction rel of blk2, resulting in the addition of a relation (rel, blk1, blk2) in the set of spatial relations. This action is defined as a sequential composition of the actions \texttt{assign\_head(blk2), move\_head(rel), keep\_at\_head(blk1)} (blk1 is a one hot tensor for the corresponding block).
\end{enumerate}
\textbf{Techniques and heuristic for efficient planning.}
Since the action space for the planner could be $o(n^2)$, where n is the number of objects we adopt the following techniques to make planning scalable/efficient:

\begin{enumerate}[noitemsep]
    \item Heuristic - We define the heuristic value h(s) for a state s, as the number of relations that are present in the goal but are absent in the scene graph corresponding to the state s. Even though this heuristic is not admissible (as it may over overestimate the cost to goal), it was found to work optimally in most of the cases.
    \item Greedy-pruning - We assume that all the actions resulting in states with higher or same heuristic value would be of the form \texttt{place-random(blk)}. This means among the actions of the form \texttt{move(rel, blk1, blk2)} we only select those that lead to states with decreased heuristic value. 
    \item Relevant-object-set - Suppose O is the set of objects that are part of atleast one of the predicate in goal. We define O' as the transitive closure of O with respect to the relation $Related$ in the initial state $s_{i}$, where $SG(s_{i})$ is scene graph for the initial state 
    \begin{equation}
    Related(a, b, s_{i}) \iff \exists dir ((dir, a, b)\in SG(s_{i}) \vee (dir, b, a) \in SG(s_{i})) \; 
    \end{equation}
    We assume O' is the relevant set of object for completing the task and actions that move any other object should not be taken.
\end{enumerate}

\section{Broader Impact}
This work creates foundational knowledge in understanding human-like spatial abstractions. This work contributes towards the development of explainable and interpret-able learning architectures that may eventually contribute towards the development of  embodied agents collaborating with and assisting humans in performing tasks. No negative impact of this work is envisioned.

\section{Hyperparameters, Architecture details and Ground Truth Concepts}
\subsection{Architecture for neural modules}
\textbf{Action Simulator:}
\begin{lstlisting}[language=Python, caption=Action Simulator Network in PyTorch]
import torch.nn as nn

class ActionSimulatorNetwork(nn.Module):
    def __init__(self, bbox_mode, hidden_size = 256):
        super(ActionSimulatorNetwork, self).__init__()
        self.bbox_mode = bbox_mode
        self.hidden_size = hidden_size

        self.action_semantics_encoder = nn.Sequential(
            nn.Linear(5, hidden_size),
            nn.ReLU(),
            nn.Linear(hidden_size, hidden_size),
            nn.ReLU()
        )
        self.argument_encoder = nn.Sequential(
            nn.Linear(5, hidden_size),
            nn.ReLU(),
            nn.Linear(hidden_size, hidden_size),
            nn.ReLU()
        )
        self.decoder = nn.Sequential(
            nn.Linear(hidden_size, hidden_size),
            nn.ReLU(),
            nn.Linear(hidden_size, 5),
            nn.Tanh()
        )
\end{lstlisting}
\textbf{Reactive Policy($\pi_{\text{neural}}$):}
\begin{lstlisting}[language=Python, caption=Neural Search in PyTorch]
import torch.nn as nn

class NeuralSearch(nn.Module):
    def __init__(self, action_space=6):
        super(NeuralSearch, self).__init__()
        self.action_space = action_space
        self.fc1 = nn.Linear(10, 256)
        # self.bn1 = nn.BatchNorm1d(256)
        self.bn1 = nn.Identity()
        self.fc2 = nn.Linear(256, 256)
        # self.bn2 = nn.BatchNorm1d(256)
        self.bn2 = nn.Identity()
        self.fc3 = nn.Linear(256, 256)
        # self.bn3 = nn.BatchNorm1d(256)
        self.bn3 = nn.Identity()
        self.fc4 = nn.Linear(256, action_space)

\end{lstlisting}

\textbf{Random Position predictor} (for grounding of \texttt{place-random(blk)}):

\begin{lstlisting}[language=Python, caption=VAE in PyTorch]
import torch.nn as nn

class VAE(nn.Module):
    def __init__(self, input_dim, latent_dim):
        super(VAE, self).__init__()
        self.input_dim = input_dim
        self.latent_dim = latent_dim
        # Encoder
        self.fc1 = nn.Linear(input_dim, 512)
        self.bn1 = nn.BatchNorm1d(512)
        self.fc2 = nn.Linear(512, 512)
        self.bn2 = nn.BatchNorm1d(512)
        self.fc3 = nn.Linear(512, 512)
        self.bn3 = nn.BatchNorm1d(512)
        self.fc4 = nn.Linear(512, 512)
        self.bn4 = nn.BatchNorm1d(512)
        self.fc51 = nn.Linear(512, latent_dim)  # Mean of the latent space
        self.fc52 = nn.Linear(512, latent_dim)  # Log-variance of the latent space (log-var for numerical stability)  
        
        # Decoder
        self.fc5 = nn.Linear(latent_dim, 512)
        self.bn5 = nn.BatchNorm1d(512)
        self.fc6 = nn.Linear(512, 512)
        self.bn6 = nn.BatchNorm1d(512)
        self.fc7 = nn.Linear(512, 512)
        self.bn7 = nn.BatchNorm1d(512)
        self.fc8 = nn.Linear(512, 512)
        self.bn8 = nn.BatchNorm1d(512)
        self.fc9 = nn.Linear(512, input_dim)
\end{lstlisting}

\textbf{On-table classifier} (for grounding of \texttt{will-not-be-floating(pred\_loc)}:

\begin{lstlisting}[language=Python, caption=Table Classifier in PyTorch]
import torch.nn as nn

class TableClassifier(nn.Module):
    def __init__(self):
        super(TableClassifier, self).__init__()
        self.fc1 = nn.Linear(5, 16)
        self.bn1 = nn.BatchNorm1d(16)
        self.fc2 = nn.Linear(16, 16)
        self.bn2 = nn.BatchNorm1d(16)
        self.fc3 = nn.Linear(16, 16)
        self.bn3 = nn.BatchNorm1d(16)
        self.fc4 = nn.Linear(16, 1)
        self.bn4 = nn.BatchNorm1d(1)
        self.sigmoid = nn.Sigmoid()

\end{lstlisting}

\subsection{Hyperparameters used in experiment}
As indicated in ~\ref{search-gen-details} for the purpose of generalization through multiple candidate plans (from 1 demonstration) we chose the top-k plans (as measured by overall IoU achieved). The k chosen for all our experiments involving MCTS was 5. (The performance of our best approach was found to be the same for k=5 to 20). For every plan we obtain 3 programs from GPT-4 by re-prompting it 3 times with the same input prompt (with temperature > 0). From the pool of these 3*k programs we chose the one with highest IoU reward by running each of them on the given demonstration. The discount factor kept for our search is $\gamma$ = 0.95, and unless explicitly specified the number of expansions steps used = 5000.

\subsection{Ground-Truth Inductive Concepts}
\label{ic-gt}
\begin{lstlisting}[language=Python, caption={Definition of inductive concepts}, label={def:inductive_concept_definition}]
######
# row
def row(length, objects):
    for i in range(length):
        keep_at_head(obj = objects)
        move_head(dir = "right")

######
# tower
def tower(height, objects):
    for i in range(height):
        keep_at_head(obj = objects)
        move_head(dir = 'top')

######
# column
def column(size, objects):
    for _ in range(size):
        keep_at_head(obj = objects)
        move_head(dir = 'front')

######
# staircase
def staircase(steps, objects):
    for step in range(1, steps+1):
        tower(height = step, objects = objects)
        move_head(dir = 'right')

######
# inverted_row
def inverted_row(num, objects):
    for i in range(num):
        keep_at_head(obj=objects)
        move_head(dir='left')

######
# inverted_column
def inverted_column(size, objects):
    for _ in range(size):
        keep_at_head(obj = objects)
        move_head(dir = 'back')
    return None

######
# inverted_staircase
def inverted_staircase(steps, objects):
    for step in range(1, steps+1):
        tower(height = step, objects = objects)
        move_head(dir = "left")

######
# diagonal_135
def diagonal_135(length, objects):
    for i in range(length):
        keep_at_head(obj = objects)
        move_head(dir = 'front')
        move_head(dir = 'left')
    return
    
######
# diagonal_315
def diagonal_315(length, objects):
    for i in range(length):
        keep_at_head(obj = objects)
        move_head(dir = 'back')
        move_head(dir = 'right')
    return

######
# diagonal_225
def diagonal_225(length, objects):
    for _ in range(length):
        keep_at_head(obj = objects)
        move_head(dir = 'back')
        move_head(dir = 'left')

######
# diagonal_45
def diagonal_45(length, objects):
    for _ in range(length):
        keep_at_head(obj = objects)
        move_head(dir = 'front')
        move_head(dir = 'right')

######
# boundary
def boundary(size, objects):
    row(length=size-1, objects=objects)
    for _ in range(size-1):
        move_head(dir = 'right')
    move_head(dir = 'front')

    column(length=size-1, objects=objects)
    for _ in range(size-1):
        move_head(dir = 'front')
    move_head(dir = 'left')

    inverted_row(length=size-1, objects=objects)
    for _ in range(size-1):
        move_head(dir = 'left')
    move_head(dir = 'back')

    inverted_column(length=size-1, objects=objects)
    for _ in range(size-1):
        move_head(dir = 'back')
    move_head(dir = 'right')
    
######
# arch_bridge
def arch_bridge(height, objects):
    staircase(steps = height, objects = objects)
    move_head(dir = 'left')
    inverted_staircase(steps = height, objects = objects)
    return

######
# x-shaped structure
def x(size, objects):
    diagonal_45(length = size, objects = objects)
    move_head(dir = 'back')
    diagonal_315(length = size, objects = objects)
    move_head(dir = 'left')
    diagonal_225(length = size, objects = objects)
    move_head(dir = 'front')
    diagonal_135(length = size, objects = objects)

######
# pyramid
def pyramid(height, objects):
    for i in range(height):
        row_length = (height * 2) - (i * 2) - 1
        row(length = row_length, objects = objects)
        if i != height - 1:
            move_head(dir = 'top')
            move_head(dir = 'right')
######
\end{lstlisting}

\section{Computational Requirements: Details}
All our experiments were run on a server with the following machine specifications.

\textbf{CPU Specification:}

\begin{longtable}{|l|l|}
\hline
\textbf{Specification} & \textbf{Value} \\ \hline
Architecture & x86\_64 \\ \hline
CPU op-mode(s) & 32-bit, 64-bit \\ \hline
Address sizes & 46 bits physical, 57 bits virtual \\ \hline
Byte Order & Little Endian \\ \hline
CPU(s) & 112 \\ \hline
On-line CPU(s) list & 0-111 \\ \hline
Vendor ID & GenuineIntel \\ \hline
Model name & Intel(R) Xeon(R) Gold 6330 CPU @ 2.00GHz \\ \hline
CPU family & 6 \\ \hline
Model & 106 \\ \hline
Thread(s) per core & 2 \\ \hline
Core(s) per socket & 28 \\ \hline
Socket(s) & 2 \\ \hline
Stepping & 6 \\ \hline
CPU max MHz & 3100.0000 \\ \hline
CPU min MHz & 800.0000 \\ \hline
BogoMIPS & 4000.00 \\ \hline
\end{longtable}

\textbf{GPU Specification:}

\begin{longtable}{|l|l|}
\hline
\textbf{Specification} & \textbf{Value} \\ \hline
\multicolumn{2}{|c|}{\textbf{GPU 1}} \\ \hline
Description & VGA compatible controller \\ \hline
Product & Integrated Matrox G200eW3 Graphics Controller \\ \hline
Vendor & Matrox Electronics Systems Ltd. \\ \hline
Physical ID & 0 \\ \hline
Bus Info & pci@0000:03:00.0 \\ \hline
Logical Name & /dev/fb0 \\ \hline
Version & 04 \\ \hline
Width & 32 bits \\ \hline
Clock & 66MHz \\ \hline
Capabilities & pm vga\_controller bus\_master cap\_list rom fb \\ \hline
Configuration & depth=32 driver=mgag200 mingnt=16  \\ \hline
Resources & irq:16 memory:91000000-91ffffff memory:92808000-9280bfff \\
          & memory:92000000-927fffff memory:c0000-dffff \\ \hline

\multicolumn{2}{|c|}{\textbf{GPU 2}} \\ \hline
Description & 3D controller \\ \hline
Product & GA102GL [A40] \\ \hline
Vendor & NVIDIA Corporation \\ \hline
Physical ID & 0 \\ \hline
Bus Info & pci@0000:17:00.0 \\ \hline
Version & a1 \\ \hline
Width & 64 bits \\ \hline
Clock & 33MHz \\ \hline
Capabilities & pm bus\_master cap\_list \\ \hline
Configuration & driver=nvidia latency=0 \\ \hline
Resources & iomemory:21000-20fff iomemory:21200-211ff irq:18 \\
          & memory:9c000000-9cffffff memory:210000000000-210fffffffff \\
          & memory:212000000000-212001ffffff memory:9d000000-9d7fffff \\
          & memory:211000000000-211fffffffff memory:212002000000-212041ffffff \\ \hline

\multicolumn{2}{|c|}{\textbf{GPU 3}} \\ \hline
Description & 3D controller \\ \hline
Product & GA102GL [A40] \\ \hline
Vendor & NVIDIA Corporation \\ \hline
Physical ID & 0 \\ \hline
Bus Info & pci@0000:ca:00.0 \\ \hline
Version & a1 \\ \hline
Width & 64 bits \\ \hline
Clock & 33MHz \\ \hline
Capabilities & pm bus\_master cap\_list \\ \hline
Configuration & driver=nvidia latency=0 \\ \hline
Resources & iomemory:28000-27fff iomemory:28200-281ff irq:18 \\
          & memory:e7000000-e7ffffff memory:280000000000-280fffffffff \\
          & memory:282000000000-282001ffffff memory:e8000000-e87fffff \\
          & memory:281000000000-281fffffffff memory:282002000000-282041ffffff \\ \hline
\end{longtable}

\textbf{Time Required: }
The time required for pretraining phase of all the neural modules is around 36 hours. For learning of inductive concepts the time taken varies from 5 minutes to 1 day depending on the search method used and the specific set of hyperparameters. However for our best approach we get the maximum performance in approx 12 minutes. Time taken for our approach during inference is less than 2 minutes per dataset.


\end{document}